\documentclass{article}

\usepackage{microtype}
\usepackage{graphicx}
\usepackage{booktabs} 
\usepackage{hyperref}

\usepackage[accepted]{icml2023}

\usepackage{amsmath}
\usepackage{amssymb}
\usepackage{mathtools}
\usepackage{amsthm}

\usepackage[capitalize,noabbrev]{cleveref}

\theoremstyle{plain}

\theoremstyle{definition}

\theoremstyle{remark}

\usepackage{amsmath,amsfonts,bm}

\newcommand{\relu}[1]{\textrm{ReLU} #1}
\newcommand{\sdmm}[1]{\texttt{SpMM} #1}
\newcommand{\mm}[1]{\texttt{MatMul} #1}
\newcommand{\sdmmean}[1]{\texttt{SpMM\_MEAN} #1}

\newcommand{\rsc}[1]{\texttt{RSC} #1}

\newcommand{\srank}[1]{\texttt{srank} #1}

\def\eqref#1{equation~(\ref{#1})}
\def\Eqref#1{Equation~(\ref{#1})}

\def\1{\bm{1}}

\def\mA{{\bm{A}}}

\def\mD{{\bm{D}}}

\def\mH{{\bm{H}}}
\def\mI{{\bm{I}}}
\def\mJ{{\bm{J}}}

\def\mW{{\bm{W}}}
\def\mX{{\bm{X}}}
\def\mY{{\bm{Y}}}

\def\mTheta{{\bm{\Theta}}}

\DeclareMathAlphabet{\mathsfit}{\encodingdefault}{\sfdefault}{m}{sl}
\SetMathAlphabet{\mathsfit}{bold}{\encodingdefault}{\sfdefault}{bx}{n}

\def\gE{{\mathcal{E}}}

\def\gG{{\mathcal{G}}}

\def\gV{{\mathcal{V}}}

\newcommand{\E}{\mathbb{E}}

\newcommand{\R}{\mathbb{R}}

\DeclareMathOperator*{\argmin}{arg\,min}

\usepackage{subcaption}
\usepackage{bbm}
\usepackage{pifont}
\usepackage{enumitem}
\usepackage{thm-restate}
\usepackage{multicol, multirow}
\definecolor{silver}{rgb}{0.852,0.852,0.852}
\newcommand{\silvercolor}{\textcolor[rgb]{0.502,0.502,0.502}}
\newcommand{\cmark}{\ding{51}}%
\newcommand{\xmark}{\ding{55}}%

\usepackage[textsize=tiny]{todonotes}

\icmltitlerunning{RSC: Accelerate Graph Neural Networks Training via Randomized Sparse Computations}

\begin{document}

\twocolumn[
\icmltitle{RSC: Accelerate Graph Neural Networks Training \\ via Randomized Sparse Computations}



\icmlsetsymbol{equal}{*}

\begin{icmlauthorlist}
\icmlauthor{Zirui Liu}{yyy}
\icmlauthor{Shengyuan Chen}{comp}
\icmlauthor{Kaixiong Zhou}{yyy}
\icmlauthor{Daochen Zha}{yyy}
\icmlauthor{Xiao Huang}{comp}
\icmlauthor{Xia Hu}{yyy}
\end{icmlauthorlist}

\icmlaffiliation{yyy}{Department of Computer Science, Rice University, Houston, TX, USA}
\icmlaffiliation{comp}{Department of Computing, The Hong Kong Polytechnic
University, Hung Hom, Hong Kong SAR}

\icmlcorrespondingauthor{Xia  Hu}{xia.hu@rice.edu}

\icmlkeywords{Machine Learning, ICML}

\vskip 0.3in
]



\printAffiliationsAndNotice{}  
\captionsetup{labelfont={it}}

\begin{abstract}

Training graph neural networks (GNNs) is extremely time-consuming because sparse graph-based operations are hard to be accelerated by community hardware.
Prior art successfully reduces the computation cost of dense matrix based operations (e.g., convolution and linear) via sampling-based approximation.
However, unlike dense matrices, sparse matrices are stored in an irregular data format such that each row/column may have a different number of non-zero entries.
Thus, compared to the dense counterpart, approximating sparse operations has two unique challenges
\textbf{(1)} we cannot directly control the efficiency of approximated sparse operation since the computation is only executed on non-zero entries;
\textbf{(2)} sampling sparse matrices is much more inefficient due to the irregular data format.
To address the issues, our key idea is to control the accuracy-efficiency trade-off by optimizing computation resource allocation layer-wisely and epoch-wisely.
For the \textbf{first} challenge, we customize the computation resource to different sparse operations, while limiting the total used resource below a certain budget.
For the \textbf{second} challenge, we cache previously sampled sparse matrices to reduce the epoch-wise sampling overhead.
To this end, we propose \textbf{R}andomized \textbf{S}parse \textbf{C}omputation.
In practice, \rsc can achieve up to $11.6\times$ speedup for a single sparse operation and $1.6\times$ end-to-end wall-clock time speedup with almost no accuracy drop.
Codes are available at \url{https://github.com/warai-0toko/RSC-ICML}.
\end{abstract}

\section{Introductions}
\begin{figure}[h!]
    \centering
        \vspace{-.5em}
    \includegraphics[width=0.99\linewidth]{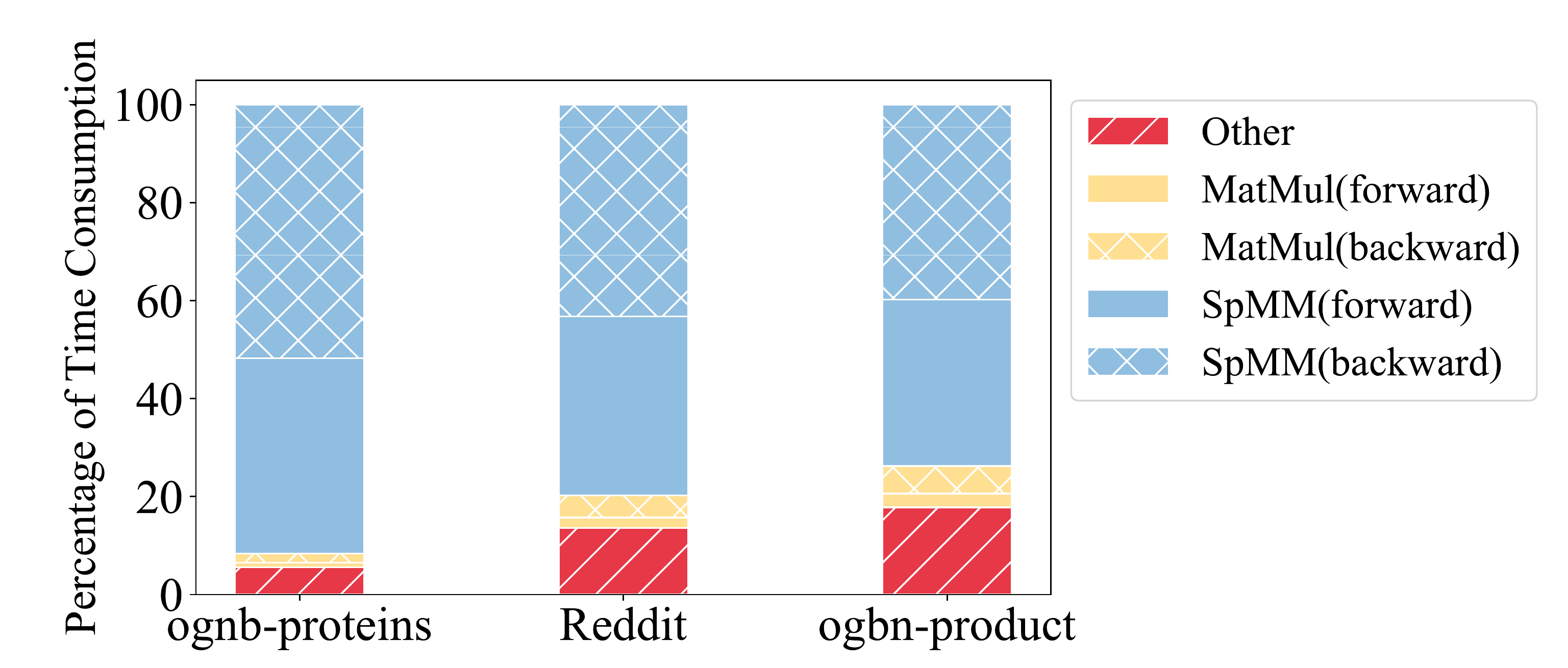}
        \vspace{-.5em}
    \caption{The time profiling of a two-layer GCNs on different datasets.
    \sdmm may take $70\%\sim 90\%$ of the total time.
    We measure the time on a single NVIDIA RTX3090 (24GB). The detailed software and hardware information can be found in Appendix \ref{app: exp_setting}.}
    \label{fig:time_percentage}
    \vspace{-.5em}
\end{figure}

Graph Neural Networks (GNNs) have achieved great success across different graph-related tasks \cite{gsage, ogb, ying2018graph, jiang2022fmp, DBLP:conf/ijcai/ZhouL0LCH22, zhou2023adaptive}.
However, despite its effectiveness, the training of GNNs is very time-consuming.
Specifically, GNNs are characterized by an interleaved execution that switches between the
aggregation and update phases.
Namely, in the aggregation phase, every node aggregates messages from its neighborhoods at each layer, which is implemented based on \emph{sparse matrix-based operations} \cite{pyg, dgl}.
In the update phase, each node will update its embedding based on the aggregated messages, where the update function is implemented with \emph{dense matrix-based operations} \cite{pyg, dgl}.
In Figure \ref{fig:time_percentage}, \sdmm and \mm are the sparse and dense operations in the aggregation and update phases, respectively. 
Through profiling, we found that the aggregation phase may take more than $90\%$ running time for GNN training.
This is because the sparse matrix operations in the aggregation phase have many random memory accesses and limited data reuse, which is hard to be accelerated by community hardwares (e.g., CPUs and GPUs) \cite{gnnbenchmark, han2016eie, duan2022a}.
Thus, training GNNs with large graphs is often time-inefficient.

Existing works towards this problem can be roughly divided into three categories. 
First, some works propose distributed GNNs training systems, which focus on minimizing the communication cost among hardware \cite{distdgl, ramezani2022learn, wan2022pipegcn, md2021distgnn, wan2022bns}. 
Second, another research line optimizes the memory access pattern of sparse operations via coalescing the memory access and fusing consecutive operations \cite{zhang2022understanding, huang2020ge, rahman2021fusedmm, wang2021tc}.
Third, some other works try to accelerate the training process from the optimization aspect, i.e., using fewer iterations to converge \cite{iglu, cong2020minimal, xu2021optimization, graphnorm}. 


In parallel, an orthogonal direction is to replace the expensive operations with their faster-approximated versions \cite{adelman2021faster, mcmatmul}.
The key idea is to sub-sample tensors onto low dimensional spaces and perform the original operations here. 
For example, for the linear operation between two matrices $\mathbf{A}\in\mathbb{R}^{n\times m}$ and $\mathbf{B}\in\mathbb{R}^{m\times q}$, 
we first obtain $\mathbf{A}'\in\mathbb{R}^{n\times k}$ and $\mathbf{B}'\in\mathbb{R}^{k\times q}$ ($k<m$) by picking $k$ representative columns of $\mathbf{A}$ and the corresponding rows of $\mathbf{B}$ \cite{mcmatmul}.
Then we approximate $\mathbf{A}\mathbf{B}\approx \mathbf{A'}\mathbf{B'}$.
With this procedure, the number of floating-point operations (FLOPs) and memory access are both reduced.
Based on the idea, previous work successfully accelerates the dense matrix based operations, such as convolution and linear operations \cite{adelman2021faster}.
The approximated operation can plug-and-play replace the exact operation to improve per-operation efficiency, and thus is compatible with most of the efficient training methods.
Despite the potential, this
perspective however has not been explored for the sparse operations in GNNs.

The approximation method reduces the computational complexity at the cost of giving noisy outputs. Thus, there naturally exists an \textbf{accuracy-efficiency trade-off}.
Compared to approximating dense matrix operations,
there are two unique challenges to optimizing the trade-off for approximated sparse operations.
\emph{First}, 
unlike the previous example of approximating linear operation, $k$ cannot directly control the efficiency (FLOPs) for sparse operations.
This is because, for dense matrices, each row/column has the same amount of parameters.
Thus the reduction of FLOPs in approximated dense operations is determined by the dimensions of the sub-sampled matrices (i.e., $k$).
However, in sparse operations, each row/column in the sparse adjacency matrix has different numbers of non-zero entries, and the computation is only executed on non-zero entries (i.e., irregular data format).
Thus, the reduction of FLOPs in the sparse operations is decided by the selection of representative rows/columns.
It lacks a mechanism to directly control the efficiency-accuracy trade-off for each sparse operation.
\emph{Second}, compared to the dense counterpart,
sub-sampling (i.e., slicing) the sparse matrix is much more time-consuming due to its irregular data format \cite{han2016eie, pyg}, which counteracts the acceleration from the FLOPs reduction.

To this end, we propose \textbf{R}andomized \textbf{S}parse \textbf{C}omputation, dubbed \rsc, the first approximation framework tailored for efficient GNN training.
Our core idea is to control the trade-off by optimizing the computation resource allocation at the ``global'' level.
Specifically, to tackle the first challenge,
\textbf{\textit{at the layer-wise level}},
we propose to customize the FLOPs of each sparse operation while limiting the total FLOPs under a certain budget.
The rationale behind this strategy is that each operation may have a different contribution to the model accuracy. Thus, we could to assign more computational resources to ``important'' operations under a certain budget.
More concretely, we frame it as a constraint optimization problem.
Then we propose a  greedy algorithm to solve it efficiently. 
To tackle the second challenge,
\textbf{\textit{at the epoch-wise level}},
we found that the selection of representative row/columns tends to remain similar across nearby iterations. 
Based on this finding, we develop a caching mechanism to reuse the previously sampled sparse matrix across nearby iterations to reduce per-epoch sampling time.
\textbf{\textit{Finally}}, inspired by the recent finding that the final stage of training usually needs smaller noise to help convergence \cite{li2019towards, dao2022monarch}, we propose to use approximated sparse operation during most of the training process, while switching back to the original sparse operation at the final stage.
This switching mechanism significantly reduces the accuracy drop, at the cost of slightly less speedup.
We summarize our contributions as follows:

\begin{itemize}[leftmargin=*]
    \item We accelerate the training of GNNs from a new perspective, namely, replacing the expensive sparse operations with their faster-approximated versions.

    \item 
    Instead of focusing on balancing the efficiency-accuracy trade-off at the operation level,
    we control the trade-off through optimizing resource allocation at the layer-wise and epoch-wise levels.

    \item 
    We propose a caching mechanism to reduce the cost of sampling sparse matrices by reusing previous results.


 \item Extensive experiments have demonstrated the effectiveness of the proposed method. 
Particularly, \rsc can achieve up to $11.6\times$ speedup for a single sparse operation and a $1.6\times$ end-to-end wall-clock time speedup with negligible ($\approx 0.3\%$) accuracy drop.
    
\end{itemize}

\nocite{zhong2022revisit}

\section{Background and Preliminary}

\subsection{Graph Neural Networks}
\label{sec: gnn time profile}
Let $\gG=(\gV,\gE)$ be an undirected graph with $\gV=(v_1,\cdots,v_{|\gV|})$ and $\gE=(e_1,\cdots,e_{|\gE|})$ being the set of nodes and edges, respectively.
Let $\mX\in\R^{|\gV|\times d}$ be the node feature matrix.
$\mA \in \R^{|\gV|\times|\gV|}$ is the graph adjacency matrix, where $\mA_{i,j}=1$ if $(v_i, v_j)\in \gE$ else $\mA_{i,j}=0$.
$\Tilde{\mA} = \Tilde{\mD}^{-\frac{1}{2}}(\mA+\mI)\Tilde{\mD}^{-\frac{1}{2}}$ is the normalized adjacency matrix, where $\Tilde{\mD}$ is the degree matrix of $\mA+\mI$.
GNNs recursively update the embedding of a node by aggregating embeddings of its neighbors.
For example, the forward pass of the $l^{\mathrm{th}}$ Graph Convolutional Network (GCN) layer~\cite{gcn} can be defined as:
\begin{equation}
\label{eq: gcnconv}
    \mH^{(l+1)}=\relu(\Tilde{\mA}\mH^{(l)}\mTheta^{(l)}),
\end{equation}
where $\mH^{(l)}$ is the node embedding matrix at the $l^{\mathrm{th}}$ layer and $\mH^{(0)}=\mX$. 
$\mTheta^{(l)}$ is the weight matrix of the $l^{\mathrm{th}}$ layer.

In practice, $\Tilde{\mA}$ is often stored in the sparse matrix format, e.g., compressed sparse row (CSR) \cite{pyg}.
From the implementation aspect, the computation of \Eqref{eq: gcnconv} can be described as:
\begin{align}
\label{eq: gcn_implement}
\mH^{(l+1)} &=\relu\Bigg(\sdmm\bigg(\Tilde\mA, \mm(\mH^{(l)}, \mTheta^{(l)})\bigg)\Bigg), \nonumber
\end{align}
where $\sdmm(\cdot,\cdot)$ is the Sparse-Dense Matrix Multiplication and $\mm(\cdot,\cdot)$ is the Dense Matrix Multiplication. 
Sparse operations, such as \sdmm, have many random memory accesses and limited data reuse. Thus they are much slower than the dense counterpart \cite{han2016eie, gnnbenchmark}.
To get a sense of the scale, we show in Figure \ref{fig:time_percentage} that for GCNs, \sdmm may take roughly $70\%\sim 90\%$ of the total training time.

\subsection{Fast Approximated $\mm$ with Sampling}
Let $\mX\in\mathbb{R}^{n\times m}$,  $\mY\in\mathbb{R}^{m\times q}$.
The goal is to efficiently estimate the matrix production $\mX\mY$.
Truncated Singular Value Decomposition (SVD) outputs provably optimal low-rank estimation of $\mX\mY$ \cite{adelman2021faster}.
However, SVD is almost as expensive as matrix production itself.
Instead, the sampling algorithm is proposed to approximate the matrix product $\mX\mY$
by sampling $k$ columns of $\mX$ and corresponding rows of $\mY$ to form smaller matrices, which are then multiplied as usual \cite{mcmatmul}.
This algorithm reduces the computational complexity from $\mathcal{O}(mnq)$ to $\mathcal{O}(knq)$.
Specifically, 
\begin{align}
        \mX\mY = \sum_{i=1}^{m} \mX_{:,i} \mY_{i,:} \nonumber &\approx \sum_{t=1}^{k} \frac{1}{s_{t}}\mX_{:,i_t}\mY_{i_t,:} \nonumber\\
        &=\texttt{approx}(\mX\mY),
\end{align}
where $\mX_{:,i}\in\mathbb{R}^{n\times 1}$ and $\mY_{i,:}\in\mathbb{R}^{1\times q}$ are the $i^{\mathrm{th}}$ column and row of $\mX$ and $\mY$, respectively.
\emph{In this paper,  we call $(\mX_{:,i}, \mY_{i,:})$ the $i^{\mathrm{th}}$ column-row pair.}
$k$ is the number of samples ($1\leq k\leq m$).
$\{p_i\}_{i=1}^m$ is a probability distribution over the column-row pairs.
$i_t\in\{1,\cdots m\}$ is the index of the sampled column-row pair at the $t^{\mathrm{th}}$ trial.
$s_{t}$ is the scale factor.
Theoretically, \cite{mcmatmul} shows that if we set $s_{t}=\frac{1}{kp_{i_t}}$, then we have $\E[\texttt{approx}(\mX\mY)]=\mX\mY$. 
Further, the approximation error $\E[||\mX\mY-\texttt{approx}(\mX\mY)||_F]$ is minimized when the sampling probabilities
$\{p_i\}_{i=1}^m$ are proportional to the product of the column-row Euclidean norms \cite{mcmatmul}:
\begin{equation}
\label{eq: col_row_norm}
    p_i = \frac{ ||\mX_{:,i}||_2\ ||\mY_{i,:}||_2}{\sum_{j=1}^{m} ||\mX_{:,j}||_2\ ||\mY_{j,:}||_2}.
\end{equation}

\subsubsection{Top-$k$ sampling}
\label{sec: top-k}
The above sampling-based method is originally developed for accelerating the general application of \mm \cite{mcmatmul}.
Directly applying it to neural networks may be sub-optimal since it does not consider the characteristic of neural network weights.
Based on the empirical observation that the distribution of weights remains centered around zero during training \cite{glorot2010understanding, han2015learning},
\cite{adelman2021faster} proposes a \textbf{top-$k$ sampling} algorithm:
\emph{Picking $k$ column-row pairs with the largest $\frac{ ||\mX_{:,i}||_2\ ||\mY_{i,:}||_2}{\sum_{j=1}^{m} ||\mX_{:,j}||_2\ ||\mY_{j,:}||_2}$ deterministically without scaling.}

Equivalently, it means $p_i$ of column-row pairs with the $k$-largest value in \Eqref{eq: col_row_norm} equals 1, otherwise it equals 0.
And $s_{i_t}$ is a constant 1.
Albeit without the scaling while sampling column-row pairs deterministically,
under on the assumption of zero-centered weight distribution, \cite{adelman2021faster} theoretically show that \textit{top-$k$ sampling still yields an unbiased estimation of $\mX\mY$ with minimal approximation error.}
Consequently, the top-$k$ sampling algorithm empirically shows a significantly lower accuracy drop when approximating the convolution and linear operations in the neural networks \cite{adelman2021faster}.



In the next section, we explore how to approximate the expensive sparse operation via the top-$k$ sampling.

\section{The Proposed Framework}

\begin{figure}[ht!]
    \centering
    \includegraphics[width=1.0\linewidth]{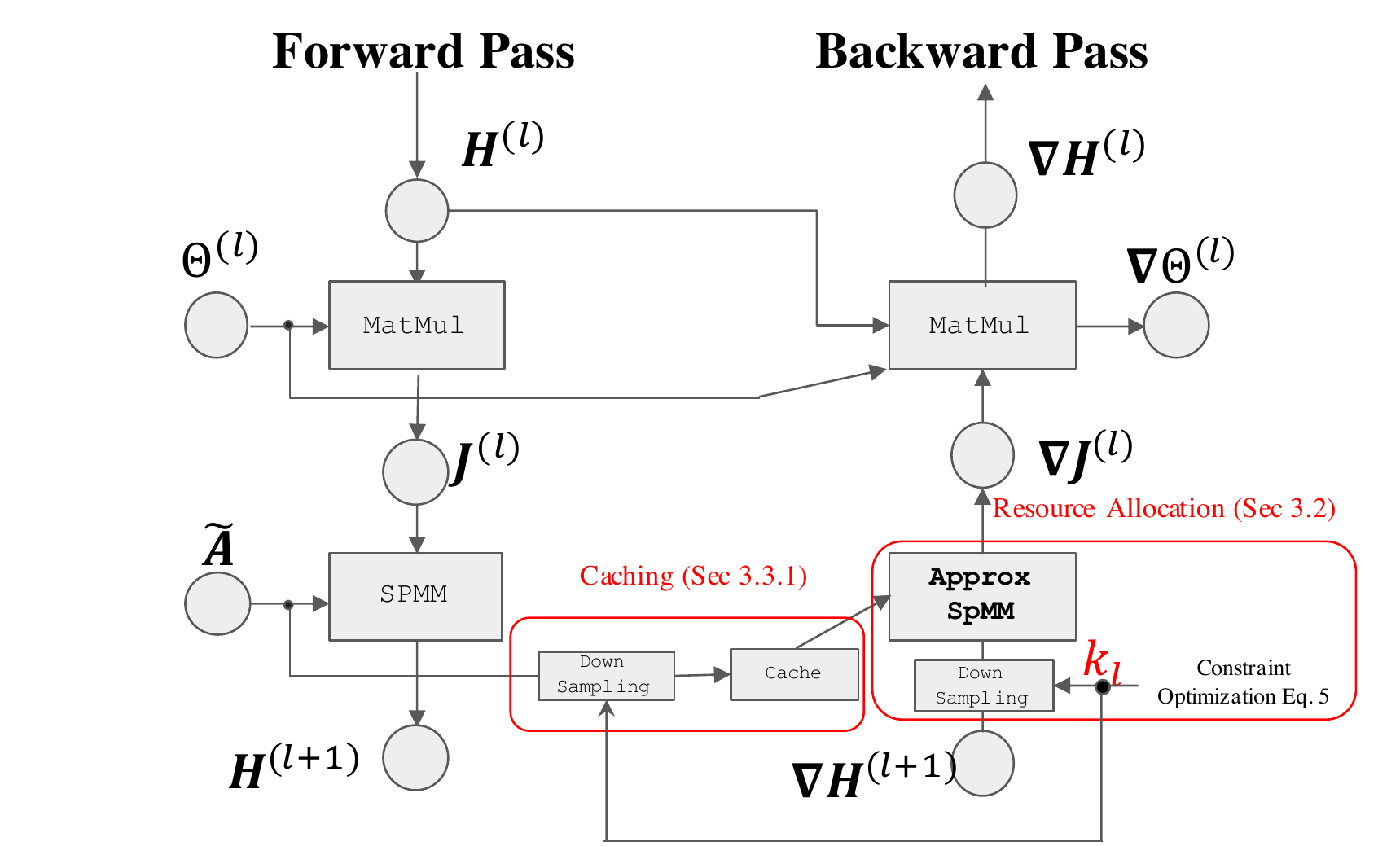}
    \caption{Overview of \rsc. 
    For convenience, ReLU is ignored. \rsc only replace the \sdmm in the backward pass with its approximated version using top-$k$ sampling (Section \ref{sec: where}).
    $k_l$ is the number of samples for top-$k$ sampling at the $l^{\mathrm{th}}$ layer, which is automatically allocated (Section \ref{sec: how}). 
    To reduce the overhead of sampling, we also cache the sampled graph and reuse it across nearby iterations (Section \ref{sec: when}).}
    \label{fig:overview}
    \vspace{-.5em}
\end{figure}

The overview of \rsc is shown in Figure \ref{fig:overview}, where we use the computation graph of GCN as an example.
We first explore which $\sdmm$ in the computation graph can be replaced with its approximated version (Section \ref{sec: where}).
Then since GNNs have multiple $\sdmm$ and each of them may have different importance to the model performance, we then automatically allocate computation resources to different $\sdmm$ (Section \ref{sec: how}).
Finally, we explore two simple and effective tricks for improving \rsc, including a caching mechanism to reduce the overhead of sampling sparse matrices (Section \ref{sec: cache}) and a switching mechanism to reduce the accuracy drop (Section \ref{sec: switch}).


\subsection{Where to Apply the Approximation}
\label{sec: where}

\subsubsection{Experimental analysis}

Each sparse operation is executed twice at each training step, i.e., one in the forward pass and the other one in the backward pass.
As shown in Figure \ref{fig:overview}, here we take \sdmm in the $l^{\mathrm{th}}$ GCN layer as an example, the forward one is $\mH^{(l+1)}=\relu(\sdmm(\Tilde{\mA}, \mJ^{(l)}))$, where $\mJ^{(l)}=\mm(\mH^{(l)}, \mTheta^{(l)})$ is the intermediate node representations.
And the backward one is 
$\nabla\mJ^{(l)}=\sdmm(\Tilde{\mA}^\top, \nabla\mH^{(l+1)})$.
$\nabla\mJ^{(l)}$ and $\nabla\mH^{(l)}$ are the gradient with respect to $\mJ^{(l)}$ and $\mH^{(l)}$, respectively.

Even though the approximation method itself is statistically unbiased,
replacing the exact sparse operation with their faster-approximated versions still injects noise to the computation graph.
As we analyzed above, each $\sdmm$ is executed twice in the training step.
Below we first experimentally analyze the impact of the injected noise in the forward pass and the backward pass.
As shown in Table \ref{tab: prelim_result}, we apply top-$k$ sampling to approximate the \sdmm in the forward pass, backward pass, or both, respectively.

\begin{table}[h!]
\centering
\caption{Preliminary results on approximating \sdmm via top-$k$ sampling. 
The model is a two-layer GCN, and the dataset is Reddit.
Here we set the $k$ as $0.1|\mathcal{V}|$ across different layers.}
\vspace{-.5em}
\label{tab: prelim_result}
\begin{tabular}{cc} 
\hline
Method               & \multicolumn{1}{c}{Reddit}  \\ 
\hline
without approximation             & 95.39±0.04                  \\
only forward        & 16.45±0.39                \\
\textbf{only backward}       & \bf{95.25±0.03}                \\
forward and backward & 80.74±1.00                \\
\hline
\end{tabular}
\end{table}

From Table \ref{tab: prelim_result}, the accuracy drop is negligible if we only replace \sdmm in the backward pass.
Notably, if we apply approximation in both the forward and backward pass, the result is significantly better than only applying top-$k$ sampling in the forward pass. 
The reason is that when only applying approximation in the forward pass, some row/columns are not included in the computation graph, so intuitively these row/columns should be excluded in the backward pass.
``forward and backward'' result in Table \ref{tab: prelim_result} is built based on this intuition such that in the backward pass, we use the column-row pairs sampled in the forward pass to compute the gradient \cite{adelman2021faster}.
However, it is still not comparable to the result of applying approximation only in the backward pass.
Below we mathematically analyze the reason behind the results in Table \ref{tab: prelim_result}.

\subsubsection{Theoretical analysis}
We first analyze the case of approximating the sparse operations in the forward pass.
Namely, replacing $\sdmm(\Tilde{\mA},\mJ^{(l)})$ with $\texttt{approx}(\Tilde{\mA}\mJ^{(l)})$.
We note that we have $\E[f(x)]\neq f(\E[x])$ for any non-linear function $f(\cdot)$, e.g., $\E[x^2]\neq \E^2[x]$.
Thus, even when the approximation method gives an unbiased estimation, i.e., $\E[\texttt{approx}(\Tilde{\mA}\mJ^{(l)})]=\Tilde{\mA}\mJ^{(l)}$,
the node embeddings $\mH^{(l+1)}$ are still biased since the activation function is non-linear. 
To see this, 
\begin{align}
\E[\mH^{(l+1)}]
    &=\E[\relu(\texttt{approx}(\Tilde{\mA}\mJ^{(l)})])\nonumber\\
    &\neq \relu(\E[\texttt{approx}(\Tilde{\mA}\mJ^{(l)})])= \mH^{(l+1)}.\nonumber
\end{align}

Thus, if we apply the approximation for the \sdmm in the forward pass, the bias will be propagated layer-by-layer and cause significantly worse results.
For the case of only approximating the sparse operation in the backward pass,
we have the following proposition:

\begin{restatable}[Proof in Appendix~\ref{app: proof}]{proposition}{propunbias}
\label{prop:unbias}
If the approximation method is itself unbiased, and we only replace the $\sdmm$ in the backward pass with its approximated version, while leaving the forward one unchanged, 
then the calculated gradient is provably unbiased.
\end{restatable}

The high-level idea is that the gradient of the activation function in the backward pass is only related to the pre-activations in the forward pass, and thus is independent of the approximation error introduced in the backward pass.
Due to the page limit, we also discuss why sampling-based approximation is suitable for accelerating GNNs in Appendix \ref{app: proof}.
As suggested by our theoretical and empirical analysis, as shown in Figure \ref{fig:overview}, 
\emph{we only approximate the sparse operations in the backward pass, while leaving all other operations unchanged.}

\subsection{How to Apply the Approximation}
\label{sec: how}

\begin{figure}[h!]
    \centering
    \includegraphics[width=0.8\linewidth]{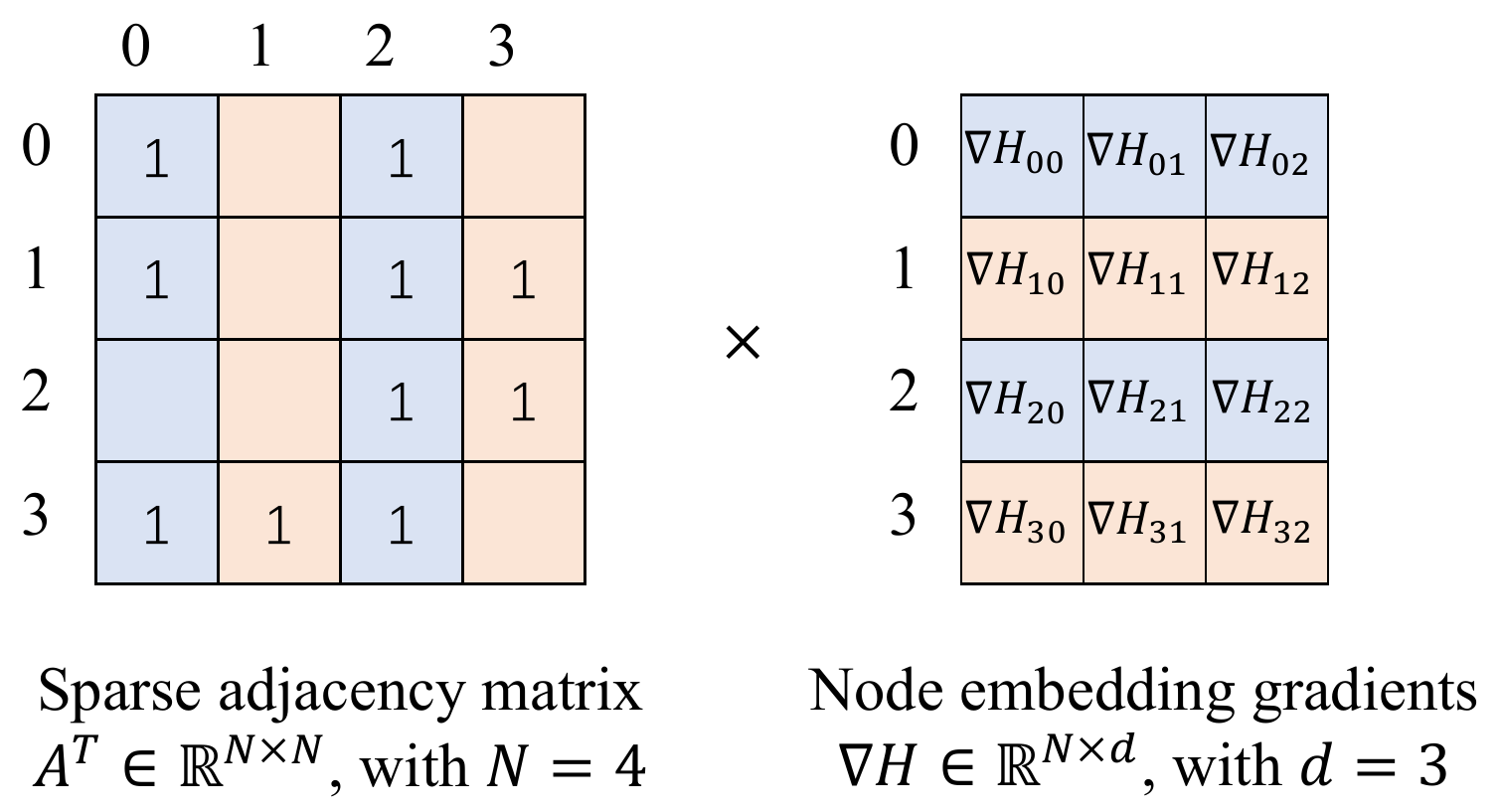}
    \vspace{-1em}
    \caption{For approximated sparse operations, the acceleration is decided by the selection of column-row pairs.}
    \vspace{-1em}
    \label{fig:selection}
\end{figure}

As we mentioned, for sparse operations, the acceleration is decided by the selection of sampled column-row pairs.
To see this, as shown in Figure \ref{fig:selection}, suppose we use top-$k$ sampling to approximate $\sdmm(\Tilde{\mA}^\top,\nabla\mH)$. 
Since the computations are only executed on the non-zero entries, so selecting the orange pairs (i.e., pair 1 and 3) will result in $\frac{3}{7}\times$ less computational cost (FLOPs) compared to selecting the blue pair (i.e., pair 0 and 2).
For both the orange and blue cases, we have $k=2$.
Thus, the number of samples $k$ cannot directly constrain the FLOPs for each individual operation.
Moreover, a GNN has multiple operations (or layers), and the model accuracy has a different sensitivity to the approximation error at different layers.
To optimize the accuracy-efficiency trade-off, our key idea is to customize the computation resources (i.e., FLOPs) for each layer by adjusting the number of samples $k_l$ in the $l$-th layer.
In this way, we minimize the impact of approximation, while limiting the overall FLOPs under a certain budget.
Based on the idea, we frame the resource allocation problem as the following constrained optimization problem:

\begin{subequations}
\label{eq: resource_allocation_opt_prob}
\begin{align}
& \min_{\left\{k_l\right\}} - \sum_{l=1}^{L}\sum_{i \in \mathrm{Top}_{k_l}}\frac{\|\Tilde\mA_{:,i}^\top\|_2 \|\nabla\mH^{(l+1)}_{i,:}\|_2}{\|\Tilde\mA\|_F\|\nabla\mH^{(l+1)}\|_F}, \label{eq: constrained opt}
\\
& s.t. \sum_{l=1}^{L} \sum_{i\in\mathrm{Top}_{k_l}} \#nnz_{i} * d_l \leq C\sum_{i=1}^{L}|\mathcal{E}| d_l, \label{eq: constraint}
\end{align}
\end{subequations}

where $C$ is the budget ($0 < C < 1$) that controls the overall reduced FLOPs.
$k_l$ is the number of samples for the top-$k$ sampling at the $l$-th layer.
$d_l$ is the hidden dimensions of $l$-th layer, and $\#nnz_{i}$ is the number of non-zero entries at the $i$-th column of $\Tilde{\mA}^\top$.
$\mathrm{Top}_{k_l}$ is the set of indices associated with the $k_l$ largest $\|\Tilde\mA_{:,i}^\top\|_2 \|\nabla\mH^{(l+1)}_{i,:}\|_2$.

\Eqref{eq: constrained opt} is equivalent to minimizing the relative approximation error $\E[\frac{||\Tilde{\mA}^\top\nabla\mH^{(l+1)}-\texttt{approx}(\Tilde{\mA}^\top\nabla\mH^{(l+1)})||_F}{\|\Tilde\mA\|_F\|\nabla\mH^{(l+1)}||_F}]$ summarized over all layers \cite{adelman2021faster}. 
Also, different sparse operations are weighted summation by the magnitude of gradient $\|\nabla\mH^{(l+1)}\|_2$, which implicitly encodes the importance of different operations.

\Eqref{eq: constraint} is the constraint that controls the overall FLOPs.
Specifically, \textbf{the FLOPs of \sdmm} between $\Tilde\mA$ and the gradient $\nabla \mH\in\mathbb{R}^{N\times d}$ is $\mathcal{O}(|\mathcal{E}|d)$ and $\sum_{j\in\mathcal{V}} \#nnz_j=|\mathcal{E}|$.
We note that \Eqref{eq: constraint} also bounds the number of memory access of \sdmm.


\subsubsection{\textbf{Greedy solution}}
\label{sec: greedy}
The above combination optimization objective is NP-hard, albeit it can be solved by dynamic programming.
However, dynamic programming is very slow, which somehow contradicts our purpose of being efficient.
Thus, \textit{we propose to use a greedy algorithm to solve it.}
Specifically, it starts with the highest $k_l=|\mathcal{V}|$ for all layers.
In each move, it chooses a $k_l$ among $\{k_l\}_{l=1}^{L}$ to reduce by a step size (e.g., $0.02|\mathcal{V}|$),
such that the increment of errors in \Eqref{eq: constrained opt} is minimal.
The greedy algorithm will stop when the current total FLOPs fits in the budget in \Eqref{eq: constraint}. 
This algorithm runs super fast, and we found that it has minimal impact on efficiency.
We provide the pseudo-code of our greedy algorithm in Algorithm \ref{algo:greedy} of Appendix \ref{app: pseudo_code}.

\begin{figure*}[t!]
    \centering
    \includegraphics[width=1\linewidth]{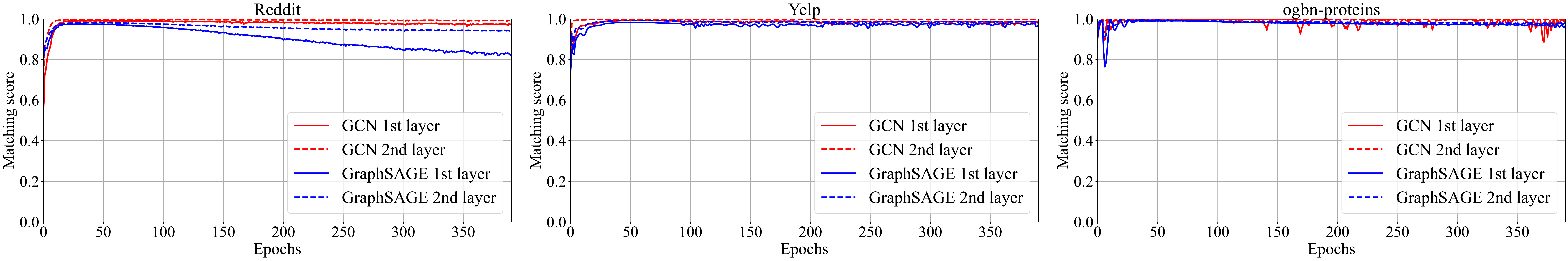}
    \vspace{-2em}
    \caption{
For each layer, the selected column-row pairs tend to be very similar across iterations. 
Models here are two-layer GCN and GraphSAGE.
Here we show the matching scores (AUC) of top-$k$ indices between every 10 steps.
}
\vspace{-1em}
    \label{fig:cache_interval10}
\end{figure*}

\begin{figure}[h!]
    \centering
    \includegraphics[width=0.8\linewidth]{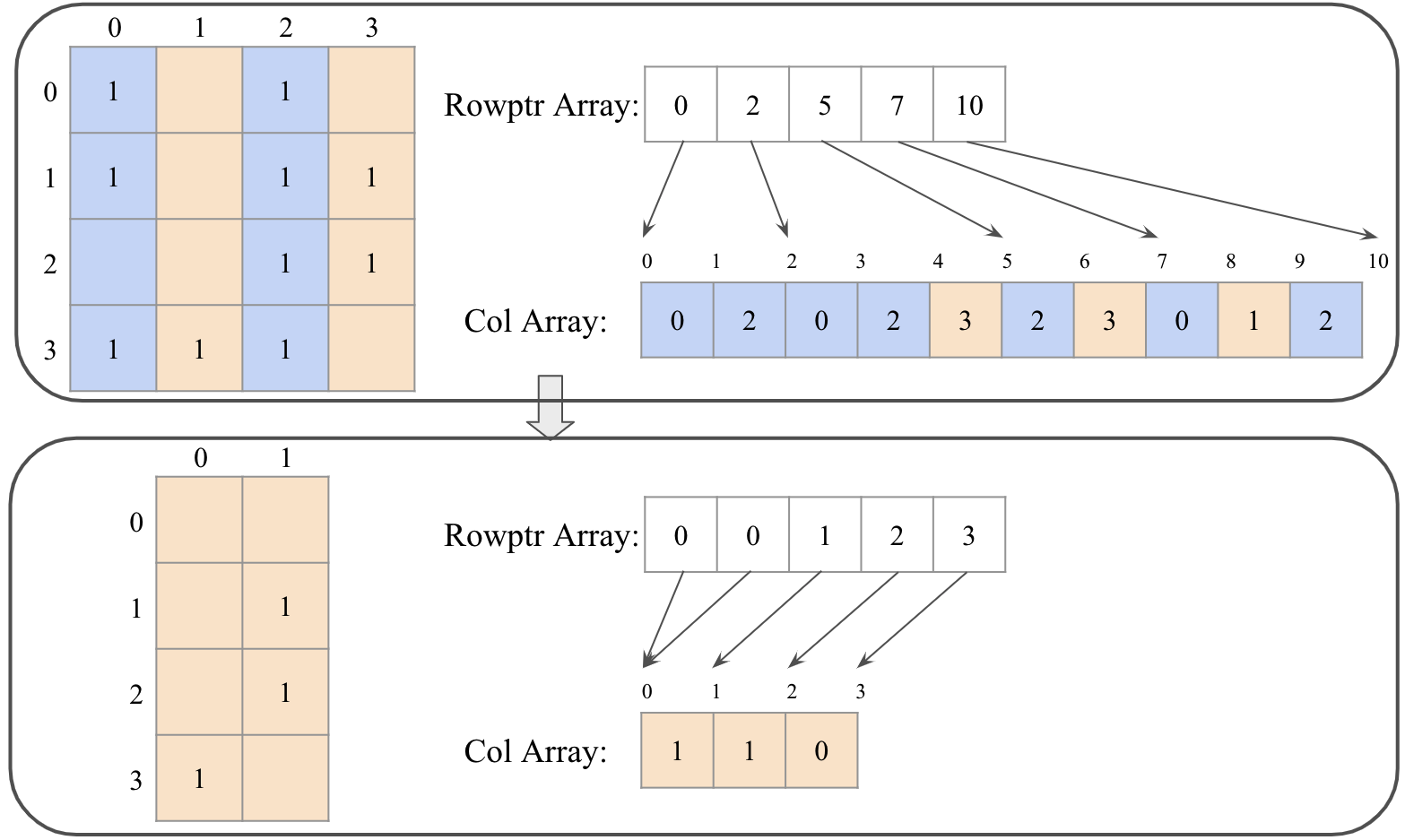}
    \vspace{-.5em}
    \caption{The process of slicing the sparse matrix in Figure \ref{fig:selection} by only reserving orange columns (in CSR format).}
    \vspace{-2em}
    \label{fig:csr_example}
\end{figure}

\subsection{When to Apply the Approximation}

\label{sec: when}
\subsubsection{\textbf{Cache the sampled sparse matrices}}
\label{sec: cache}
We first give the details about the Compressed Sparse Row (CSR) format for representing the sparse matrix here.
CSR stores nonzero values in a matrix and their position in three arrays:
index array \texttt{Rowptr}, 
column array \texttt{Col}, 
and value array \texttt{Val}. 
The elements in \texttt{Rowptr} act as the starting indices of the elements in
\texttt{Col} and \texttt{Val} that correspond to each row. 
Specifically, the elements of row $i$ are stored in indices \texttt{Rowptr}[i] to \texttt{Rowptr[i+ 1]} $-$ 1 
of \texttt{Col} and \texttt{Val} . 
The elements in \texttt{Col} and \texttt{Val} are the column index and value in that column, respectively.
Figure \ref{fig:csr_example} shows the CSR
format of the matrix shown in Figure \ref{fig:selection}. 
We ignore the \texttt{Val} array here for illustration convenience.

Executing the top-$k$ sampling contains two steps: First, it decides the indices corresponding to the top-$k$ largest column row norms in \Eqref{eq: col_row_norm}.
Second, slicing the matrices according to the indices.
In practice, the overhead of the first step can be ignored.
However, unlike dense matrices, slicing the adjacency matrix is much slower due to its irregular data format.
To see this, 
suppose the top-$k$ indices of the sparse matrix in Figure \ref{fig:selection} correspond to the orange column-row pairs.
Figure \ref{fig:csr_example} shows the process of slicing the adjacency matrix in CSR format by reserving only the orange columns.
Slicing sparse matrices requires to re-process the graph to build the new \texttt{Rowptr} and \texttt{Col} \cite{pyg}, which introduces significant time overhead, especially for large graphs.

For the full graph training, we use the same adjacency matrix across different epochs\footnote{For sub-graph based training, we can first sample all of the sub-graphs offline. 
Then during the training, we apply the caching mechanism to each sampled graph.}.
We made a crucial observation that the top-$k$ indices in the adjacency matrix tend to be the same across iterations.
In Figure \ref{fig:cache_interval10}, we plot the AUC score of top-$k$ indices between every iteration $t$ and iteration $t + 10$ for each layer throughout the whole training process.
Here we note that AUC score is a commonly used ranking measure and a 1.0 AUC score means the ranking of column-row pairs is identical across iterations.
The results in Figure \ref{fig:cache_interval10} indicate that the top-$k$ indices won’t change significantly within a few iterations.
Thus, as shown in Figure \ref{fig:overview}, we propose to reuse the sampled adjacency matrix for each layer across nearby iterations.

\noindent
\textbf{Discussion.}
The rationale behind the success of caching is the slow rate of change in the learned embeddings within GNNs \cite{gnnautoscale, wan2022bns}. 
Prior research has leveraged this ``staleness'' of embeddings to enhance the efficiency of GNN training [1, 2]. The success of caching can also be explained by the staleness: if embeddings (and their gradients) across consecutive steps remain nearly identical, the sampled sparse matrix will also exhibit minimal variation.
Later we experimentally show that the caching mechanism does not impact the model performance a lot, but leads to a significant speedup.

\subsubsection{\textbf{Switch back at the end}}
\label{sec: switch}

When training neural networks, the common practice is to use a large learning rate for
exploration and anneal to a small one for final convergence
\cite{li2019towards}.
The rationale behind this strategy is that, at the end of the training process, we need to fine-tune our model with small noise for convergence.
Since our approximation sparse operations will bring extra noise to the gradient,
intuitively, we can switch back to the original sparse operations to help convergence.
More formally,
we propose to use approximated sparse operation during most of the training process, while switching back to the original sparse operation at the final stage.
We experimentally show that this switching mechanism significantly reduces the accuracy drop at the cost of slightly less acceleration effect.

\textbf{We note that the switching mechanism is not proposed in this paper.}
The switching mechanism takes inspiration from previous work \citet{dao2022monarch}, and both our work and \citet{dao2022monarch} utilize the switching mechanism to minimize the impact of approximation.


\section{Related work and Discussion}

Due to the page limit, we first discuss the related work on approximated matrix multiplication.
Other related topics, i.e., subgraph-based training, randomized GNN training, and non-approximated GNN acceleration, can be found in Appendix \ref{app: related_work}.

\textit{\textbf{Approximated Matrix Multiplication.}}
The approximated matrix production can be roughly divided into three categories.
However, only a few of them can be used for accelerating GNN training.
Specifically,
(1) Random walk-based methods \cite{cohen1999approximating} performs random walks on a graph representation of the dense
matrices, but is only applicable to non-negative matrices;
(2) Butterfly-based methods \cite{chen2021pixelated, dao2022monarch} replace dense matrices with butterfly matrices. It is not applicable to $\sdmm$ in GNNs because the adjacency matrix often cannot be reduced to a butterfly matrix.
(3) Column-row sampling methods\cite{drineas2006fast, drineas2001fast} sample the input matrices with important rows and columns, then perform the production on the sampled matrix as usual.

\section{Limitations}

First, to guarantee the model accuracy, we only replace the sparse operation in the backward pass.
Thus the upper bound of \rsc's speedup is limited.
However, we note that the backward pass usually is more time-consuming than the forward pass, which is also empirically shown in Table \ref{tab: efficiency_operation}.
Second, some GNNs rely on the scatter-and-gather instead of $\sdmm$ (and its variant) to perform the aggregation, such as GAT \cite{gat}.
They are not covered in this paper.
However, scatter-and-gather based GNNs
can also be accelerated by \rsc because the column-row sampling is also applicable to scatter and gather operation.
Similarly,
the caching and switching mechanisms are also applicable to them.
However, for the resource allocation Algorithm \ref{algo:greedy}, the scatter and gather operations require tailored error bound and the computation cost modeling in \Eqref{eq: resource_allocation_opt_prob}.
We leave it as future work.

\section{Experiments}
\label{sec: exp}
We verify the effectiveness of our proposed framework via answering the following research questions:
\textbf{Q1}: 
    How effective is \rsc in terms of accuracy with reduced training time?
\textbf{Q2}: 
    How effective is our proposed allocation strategy compared to the uniform allocation strategy? 
\textbf{Q3}:
    What is the layer-wise ratio assigned by \rsc?
\textbf{Q4}: 
    How effective is the caching and switching mechanism in terms of the trade-off between efficiency and accuracy? 
If without explicitly mentioned, all reported results are averaged over ten random trials
\begin{table*}[h!]
\centering
\caption{Comparison on the efficiency at the operation level. fwd/bwd is the wall-clock time for a single forward/backward pass (ms). \sdmmean corresponds to the MEAN aggregator used in GraphSAGE (Appendix \ref{app: sage_analysis}).}
\vspace{-.5em}
\label{tab: efficiency_operation}
\resizebox{\linewidth}{!}{
\begin{tabular}{cccccccccc} 
\hline
                                                                             &          & \multicolumn{2}{c}{Reddit}                                        & \multicolumn{2}{c}{Yelp}                                         & \multicolumn{2}{c}{\begin{tabular}[c]{@{}c@{}}\textit{ogbn-}\\\textit{proteins}\end{tabular}} & \multicolumn{2}{c}{\begin{tabular}[c]{@{}c@{}}\textit{ogbn-}\\\textit{products}\end{tabular}}  \\
                                                                             &          & fwd    & bwd                                                      & fwd   & bwd                                                      & fwd    & bwd                                                                & fwd    & bwd                                                                 \\ 
\hline
\multirow{2}{*}{\sdmm}                                                        & Baseline & 36.28~ & 44.23~                                                   & 26.88 & 34.38                                                    & 31.72~ & 42.99                                                              & 261.03 & 316.80~                                                             \\
                                                                             & +\rsc        & -      & 3.81 (11.6$\times$)                                      & -     & ~9.86 (3.49$\times$)                                     & -      & 14.87 (2.89$\times$)                                               & -      & 35.28 (8.98$\times$)                                                \\ 
\hline
\multirow{2}{*}{\begin{tabular}[c]{@{}c@{}}\sdmmean\\(Appendix \ref{app: sage_analysis})\end{tabular}} & Baseline & 36.21~ & 44.27                                                    & 26.78 & 34.38~                                                   & 31.80~ & 43.11                                                              & 261.03 & 316.84                                                              \\
                                                                             & +\rsc       & -      & ~7.47 (\textcolor[rgb]{0.122,0.125,0.141}{5.92$\times$}) & -     & 19.62 (\textcolor[rgb]{0.122,0.125,0.141}{1.75$\times$}) & -      & 5.22~ (8.26$\times$)                                               & -      & 71.59~(4.43$\times$)                                                \\
\hline
\end{tabular}}
\vspace{-.5em}
\end{table*}

\begin{table*}[t!]
    \centering
    \caption{Comparison on the test accuracy/F1-micro/AUC and speedup on four datasets. \textbf{Bold faces} indicate the accuracy drop is negligible ($\approx 0.3\%$) or the result is better compared to the baseline.The hardware here is a RTX3090 (24GB).}
    \vspace{-.5em}
    \label{tab: perf_vs_compress}
    \resizebox{\linewidth}{!}{
    \begin{tabular}{cccccccccccccc} 
    \hline
    \multicolumn{2}{c}{\begin{tabular}[c]{@{}c@{}}\# nodes\\\# edges\end{tabular}}                              & \multicolumn{3}{c}{\begin{tabular}[c]{@{}c@{}}230K\\11.6M\end{tabular}}              & \multicolumn{3}{c}{\begin{tabular}[c]{@{}c@{}}717K\\7.9M\end{tabular}}               & \multicolumn{3}{c}{\begin{tabular}[c]{@{}c@{}}132K\\39.5M\end{tabular}}                       & \multicolumn{3}{c}{\begin{tabular}[c]{@{}c@{}}2.4M\\61.9M\end{tabular}}                        \\
    \multirow{2}{*}{Model}                                                           & \multirow{2}{*}{Methods} & \multicolumn{3}{c}{Reddit}                                                           & \multicolumn{3}{c}{Yelp}                                                             & \multicolumn{3}{c}{\begin{tabular}[c]{@{}c@{}}\textit{ogbn-}\\\textit{proteins}\end{tabular}} & \multicolumn{3}{c}{\begin{tabular}[c]{@{}c@{}}\textit{ogbn-}\\\textit{products}\end{tabular}}  \\
                                                                                     &                          & Acc.       & \begin{tabular}[c]{@{}c@{}}Budget \\$C$\end{tabular} & Speedup      & F1-micro   & \begin{tabular}[c]{@{}c@{}}Budget \\$C$\end{tabular} & Speedup      & AUC       & \begin{tabular}[c]{@{}c@{}}Budget \\$C$\end{tabular} & Speedup            & Acc.       & \begin{tabular}[c]{@{}c@{}}Budget \\$C$\end{tabular} & Speedup                \\ 
    \hline
    \multirow{2}{*}{\begin{tabular}[c]{@{}c@{}}Graph-\\SAINT\end{tabular}}           & Baseline                 & 96.40\silvercolor{±0.03} & 1                                                        & 1$\times$    & 63.30\silvercolor{±0.14} & 1                                                        & 1$\times$    & —          & —                                                           & —                  & 79.01\silvercolor{±0.21} & 1                                                        & 1$\times$              \\
                                                                                     & +\rsc                        & \bf{96.24}\silvercolor{±0.03} & 0.1                                                      & 1.11$\times$ & \bf{63.34}\silvercolor{±0.18} & 0.1                                                      & 1.09$\times$ & —          & —                                                           & —                  & \bf{78.99}\silvercolor{±0.32} & 0.3                                                      & 1.04$\times$           \\ 
    \hline
    \multirow{2}{*}{GCN}                                                             & Baseline                 & 95.33\silvercolor{±0.03} & 1                                                        & 1$\times$    & 44.28\silvercolor{±1.04} & 1                                                        & 1$\times$    & 71.99\silvercolor{±0.66} & 1                                                           & 1$\times$          & 75.74\silvercolor{±0.11} & 1                                                        & 1$\times$              \\
                                                                                     & +\rsc                        & \bf{95.13}\silvercolor{±0.05} & 0.1                                                      & 1.47$\times$ & \bf{46.09}\silvercolor{±0.54} & 0.1                                                      & 1.17$\times$ & \bf{71.60}\silvercolor{±0.45} & 0.3                                                         & 1.51$\times$       &       \bf{75.44}\silvercolor{±0.21}     &  0.3                                                        &  1.35$\times$                      \\ 
    \hline
    \multirow{2}{*}{\begin{tabular}[c]{@{}c@{}}GraphSAGE\\(full batch)\end{tabular}} & Baseline                 & 96.61\silvercolor{±0.05} & 1                                                        & 1$\times$    & 63.06\silvercolor{±0.18} & 1                                                        & 1$\times$    & 76.09\silvercolor{±0.77} & 1                                                           & 1$\times$          & 78.73 \silvercolor{± 0.12} & 1                                                        & 1$\times$              \\
                                                                                     & +\rsc                        & \bf{96.52}\silvercolor{±0.04} & 0.1                                                      & 1.32$\times$ & \bf{62.89}\silvercolor{±0.19} & 0.1                                                      & 1.13$\times$ & \bf{76.30}\silvercolor{±0.42} & 0.3                                                         & 1.60$\times$       &      \bf{78.50}\silvercolor{± 0.09}      &     0.1                                            &      1.53$\times$                          \\ 
    \hline
    \multirow{2}{*}{GCNII}                                                           & Baseline                 & 96.71\silvercolor{±0.07} & 1                                                        & 1$\times$    & 63.45\silvercolor{±0.17} & 1                                                        & 1$\times$    & 73.79\silvercolor{±1.32} & 1                                                           & 1$\times$          & —          & —                                                        & —                      \\
                                                                                     & +\rsc                        & \bf{96.50}\silvercolor{±0.12} & 0.3                                                      & 1.45$\times$ & \bf{63.57}\silvercolor{±0.21} & 0.1                                                      & 1.19$\times$ & \bf{75.20}\silvercolor{±0.54} & 0.5                                                         & 1.41$\times$       & —          & —                                                        & —                      \\
    \hline
    \end{tabular}}
    \vspace{-1em}
    \end{table*}
    
\subsection{Experimental Settings}
\label{sec: exp_setting}
\textbf{Datasets and Baselines.}
To evaluate \rsc, we adopt four common large-scale graph benchmarks from different domains, i.e., Reddit \cite{gsage}, Yelp \cite{gsaint}, \textit{ogbn-proteins} \cite{ogb}, and \textit{ogbn-products} \cite{ogb}.
We evaluate \rsc under both the mini-batch training and full-batch training settings.
For the mini-batch training setting, we integrate \rsc with one of the state-of-the-art sampling methods, GraphSAINT \cite{gsaint}. 
For the full-batch training setting, we integrate \rsc with three popular models:
two commonly used shallow models, namely, GCN \cite{gcn} and GraphSAGE \cite{gsage}, and one deep model GCNII \cite{gcn2}.
To avoid creating confusion, 
GCN, GraphSAGE, and GCNII are all trained with the whole graph at each step.
For a fair comparison, we use the MEAN aggregator for GraphSAGE and GraphSAINT throughout the paper.
Details about the hyperparameters and datasets are in Appendix \ref{app: exp_setting}.

\noindent
\textbf{Hyperparameter settings.}
\rsc contains three parts. 
First, the allocation strategy.
We choose the overall budget $C$ in \Eqref{eq: constraint} from $\{0.1, 0.3, 0.5\}$.
We run the resource allocation strategy every \textbf{ten} steps.
The step size $\alpha$ in Algorithm \ref{algo:greedy} is set as $0.02|\mathcal{V}|$.
Second, the caching mechanism.
According to Figure \ref{fig:cache_interval10}, we sample the adjacency matrix every \textbf{ten} steps and reuse the sampled matrices for nearby steps. 
Third, the switching mechanism, where we apply \rsc for \textbf{80\%} of the total epochs, while switching back to the original operations for the rest of the \textbf{20\%} epochs.
Due to the page limit,
We present a detailed hyperparameter study in Appendix \ref{app: exp_results} Figure \ref{fig: ablation_learning_curve} and Figure \ref{fig: hp}.

\noindent
\textbf{Evaluation metrics.}
To evaluate the practical usage of \rsc,
we report the wall clock time speedup measured on GPUs. Specifically, the speedup equals $T_{\text{baseline}} / T_{rsc}$, where $T_{\text{baseline}}$ and $T_{rsc}$ are the wall clock training time of baseline and \rsc, respectively.
We note that \emph{the $T_{rsc}$ includes the running time of the greedy algorithm, and the effects of caching and switching.}

\vspace{-.5em}
\subsection{Performance Analysis}
\subsubsection{Accuracy-efficiency trade-off}
To answer \textbf{Q1}, we summarize the speedup and the test accuracy/F1-micro/AUC of different methods in Table \ref{tab: perf_vs_compress}.
Since \rsc accelerates the sparse operation in the backward pass, we also provide the detailed efficiency analysis in Table \ref{tab: efficiency_operation}.
In summary, we observe:

\begin{figure*}[t!]
    \centering
    \begin{subfigure}[h]{0.3\linewidth}
    \includegraphics[width=\linewidth]{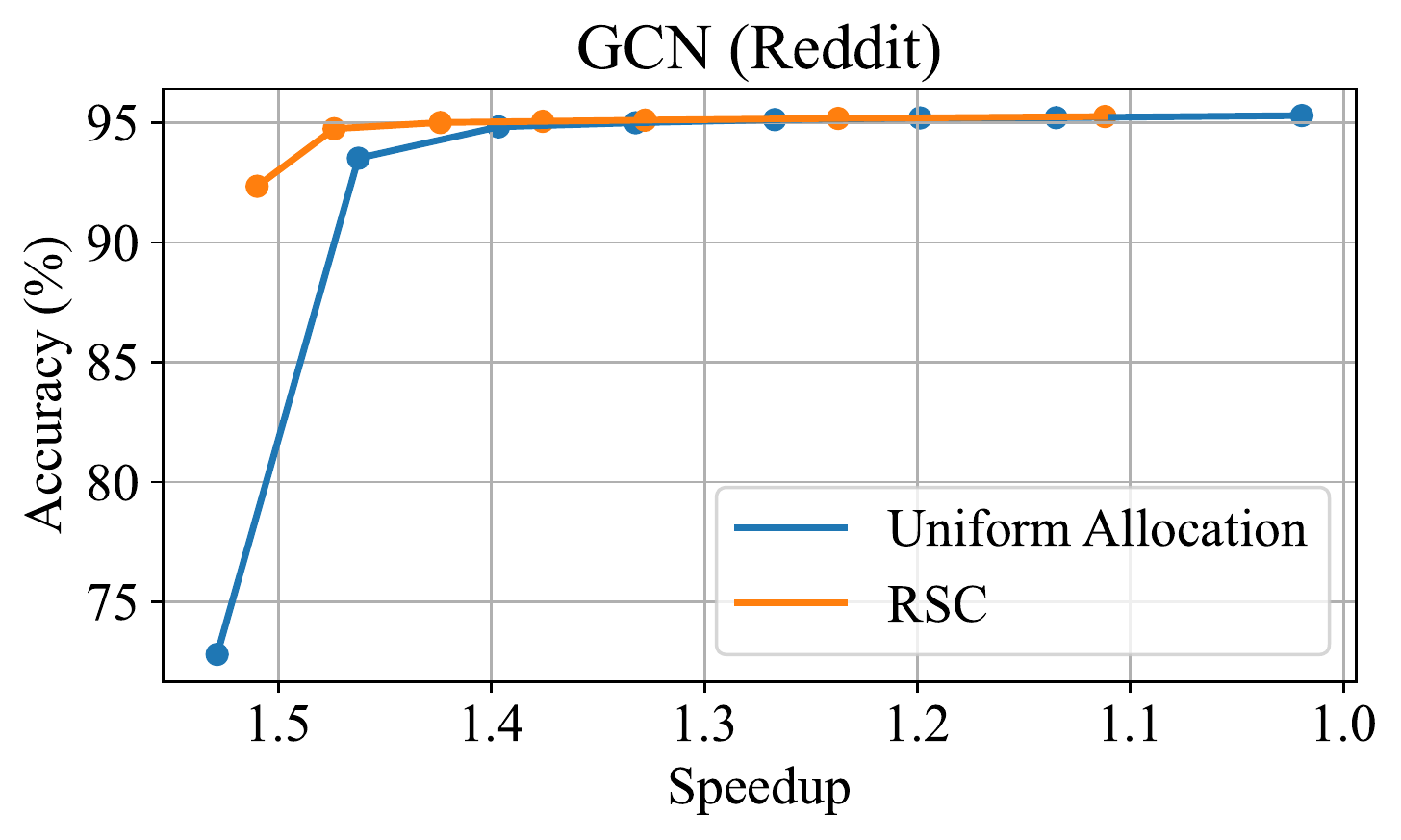}
    \end{subfigure}
    \begin{subfigure}[h]{0.3\linewidth}
    \includegraphics[width=\linewidth]{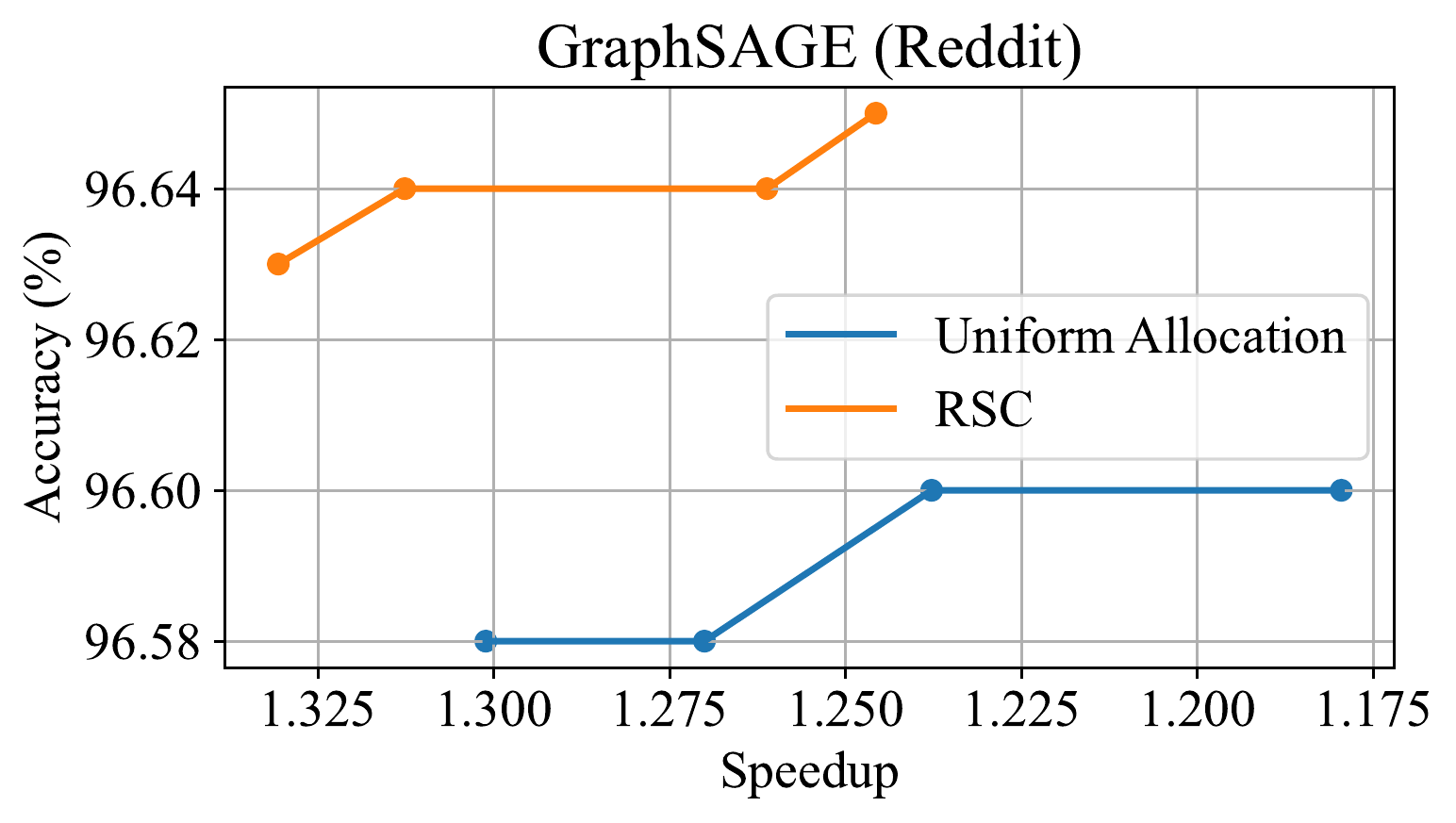}
    \end{subfigure}
    \begin{subfigure}[h]{0.3\linewidth}
    \includegraphics[width=\linewidth]{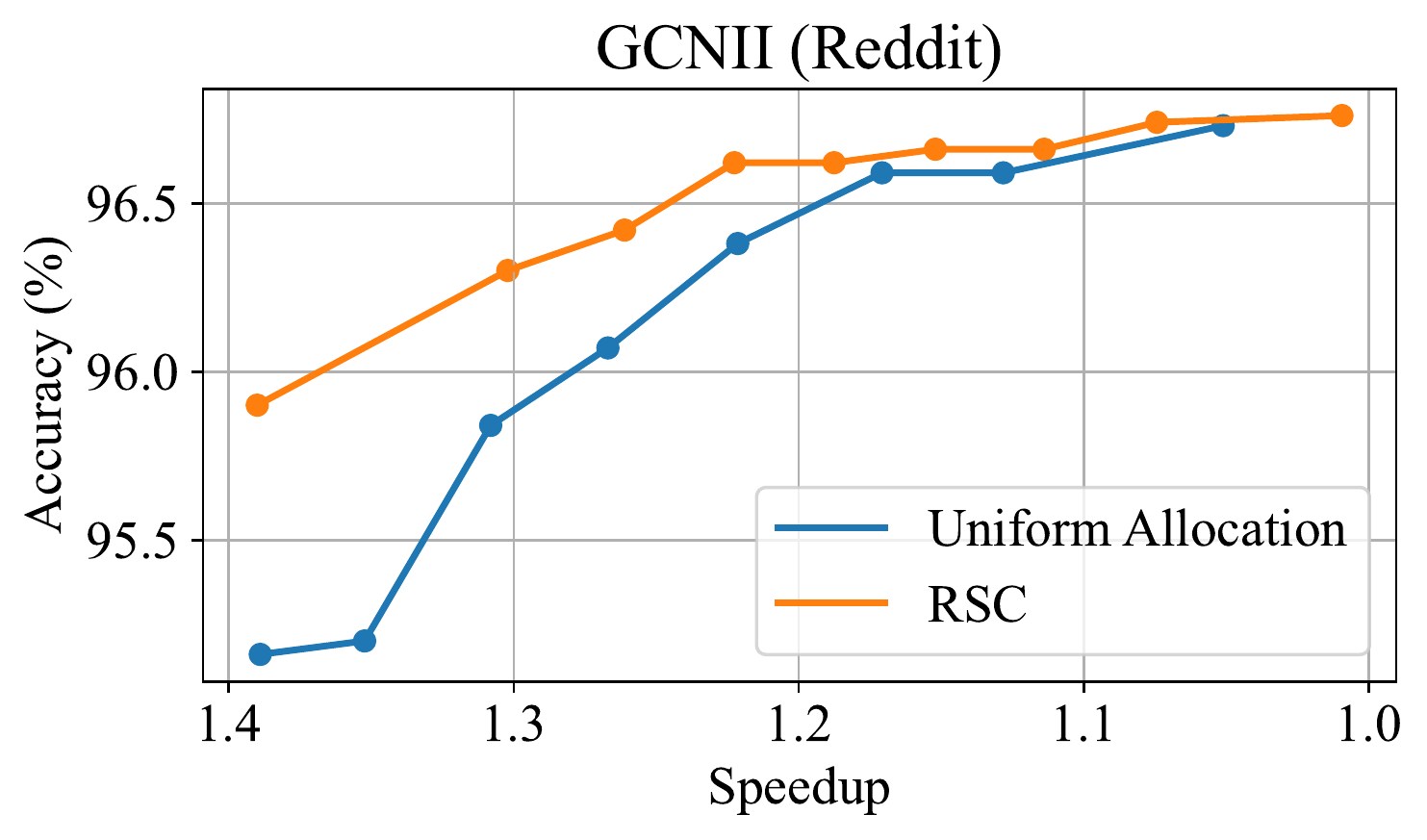}
    \end{subfigure}
        \vspace{-1em}
    \caption{The Pareto frontier of the accuracy-efficiency trade-off for \rsc and the uniform allocation.
    Here we disabled the caching and switch mechanism for a fair comparison. More results can be found in Appendix \ref{app: exp_results}}
    \vspace{-1.5em}
    \label{fig: ablation_pareto}
\end{figure*}

\begin{table}[t!]
    \centering
    \caption{Ablation on the caching and switching mechanism. Experiments are conducted on \textit{ogbn-proteins}. All results are averaged over five random trials.}
    \vspace{-.5em}
    \label{tab: ablation_cache_switch}
    \small
    \resizebox{\linewidth}{!}{
    \begin{tabular}{ccccc}
    \hline
    \multicolumn{1}{l}{Ablation on} & Caching & Switching & \multicolumn{1}{c}{AUC} & \multicolumn{1}{c}{Speedup}  \\ 
    \hline
    \multirow{4}{*}{GCN}            & \xmark   & \xmark     & 71.60 \silvercolor{± 0.66}    &    1.19$\times$     \\
                                    & \xmark   & \cmark     & 72.19 \silvercolor{± 0.79}   &     1.14$\times$     \\      
                                    & \cmark   & \xmark     & 69.80 \silvercolor{± 0.60}    &    1.60$\times$     \\
                                    & \cmark   & \cmark     & 71.60 \silvercolor{± 0.45}    &  1.51$\times$         \\ 
    \hline
    \multirow{4}{*}{GraphSAGE}      & \xmark   & \xmark     & 75.23 \silvercolor{± 0.79}    &       1.37$\times$                       \\
                                    & \xmark   & \cmark     & 76.39 \silvercolor{± 0.39}    &    1.32$\times$                         \\
                                    & \cmark   & \xmark     & 75.53 \silvercolor{± 0.60}   &  1.78$\times$                          \\
                                    & \cmark   & \cmark     & 76.30 \silvercolor{± 0.42}   &  1.60$\times$                           \\ 
    \hline
    \multirow{4}{*}{GCNII}          & \xmark   & \xmark     &    74.07 \silvercolor{± 0.83 }                &    1.10$\times$                      \\
                                    & \xmark   & \cmark     &   74.50 \silvercolor{± 0.52}                     &  1.04$\times$                        \\
                                    & \cmark   & \xmark     &     72.47 \silvercolor{± 0.75}                 &    1.46$\times$                          \\
                                    & \cmark   & \cmark     &     75.20 \silvercolor{± 0.54}                  &    1.41$\times$                       \\
    \hline
    \end{tabular}}
    \vspace{-2em}
    \end{table}
    
\textbf{\ding{182}} 
\textit{
At the operation level, \rsc can accelerate the sparse operation in the backward pass by up to $11.6\times$.
For end-to-end training,
the accuracy drop of applying \rsc
over baselines is negligible (0.3\%) across different models and datasets, while achieving up to $1.6\times$ end-to-end wall clock time speedup.}
The gap between the operation speedup and the end-to-end speedup is due to the following two reasons.
\emph{First}, we focus on accelerating the sparse computations in GNNs, which is the unique bottleneck to GNNs.
The other dense computations can certainly be accelerated by approximation methods, but this is beyond the scope of this paper.
\emph{Second}, we only accelerate the sparse computation in the backward pass instead of the forward one to guarantee performance.
We note that for approximation methods that accelerate the training process at operation level, a $1.2\approx1.3\times$ wall-clock speedup with negligible accuracy drop can be regarded as non-trivial (for details, please see Table 1 in \cite{adelman2021faster}),
especially considering that these approximation methods are orthogonal to most of the existing efficient training methods.
For GraphSAINT, the speedup of \rsc is around $1.1\times$, which is smaller than the full graph training.
This is because for subgraph-based training, the equivalent ``batch size'' is much smaller than the full graph counterparts.
As a result, the GPU utility is low since it does not assign each processor a sufficient amount of work and the bottleneck is the mini-batch transfer time  \cite{kaler2022accelerating}.
We note that the mini-batch sampling and transfer time can be optimized from the system perspective \cite{kaler2022accelerating}, which is orthogonal to our work. The speedup is expected to be larger when the mini-batch sampling time is optimized.



\subsubsection{Ablation on resource allocation.}
\label{sec: ablation_on_resource_allocation}

Due to the page limit, we first show the running time of the greedy algorithm in Appendix \ref{app: exp_results} Table \ref{tab: greedy_runtime}.
We conclude that the overhead of the greedy algorithm is negligible compared to the acceleration effect of \rsc.
To answer \textbf{Q2}, we compare \rsc with the uniform allocation strategy, i.e., setting $k_l=C|\mathcal{V}|$ for all sparse operations in the backward pass.
As shown in Figure \ref{fig: ablation_pareto}, we plot the Pareto frontier of the accuracy-efficiency trade-off on the Reddit dataset for \rsc and the uniform strategy with different $C$. 
For a fair comparison, we disabled the caching and switching mechanism.
Due to page limit, more results are shown in Appendix \ref{app: exp_results}.
We observe that:
\textbf{\ding{183}} 
\textit{\rsc exhibits a superior trade-off between accuracy and efficiency compared to the uniform allocation, especially under high speedup regime.}
Namely, compared to the uniform allocation, \rsc can achieve higher model accuracy under the same speedup.
This can be explained by the fact that each operation has a different importance to the model performance.
\rsc can automatically allocate more resources to important operations under a given total budget.

To answer \textbf{Q3}, due to the page limit, we visualize the allocated $k_l$ for each layer across iterations in Appendix \ref{app: exp_results} Figure \ref{fig: ablation_layer-wise_ratio}, and the degree of picked nodes in Appendix \ref{app: exp_results} Figure \ref{fig: ablation_node_deg}.
We observe:
\textbf{\ding{184}} 
\textit{The $k_l$ assigned by \rsc evolves along with the training.}


\subsubsection{Ablation on caching and switching.}
In section \ref{sec: ablation_on_resource_allocation}, we have shown the superior results of the proposed resource allocation strategy.
As we mentioned in Section \ref{sec: when}, we also introduce two simple tricks to for improving \rsc, i.e., the caching and switching mechanism.
To verify the effect of each of them (\textbf{Q4}), we conduct incremental evaluations on GCN, GraphSAGE and GCNII with \textit{ogbn-proteins}, which are summarized in Table \ref{tab: ablation_cache_switch}.
The row without caching and switching in Table \ref{tab: ablation_cache_switch} corresponds to the results with the proposed resource allocation strategy.
We observe:
\textbf{\ding{185}} 
\textit{Switching mechanism significantly improves the model performance, at the cost of slightly less acceleration effect.}
As we analyzed in Section \ref{sec: switch}, the improvement can be explained by the fact that the final training stage requires smaller gradient noise to help convergence.
\textbf{\ding{186}} 
\textit{Caching mechanism significantly improves the wall-clock time speedup, at the cost of worse model performance.}
Although caching mechanism can reduce the overhead of sampling, the performance drop is too large ($>1\%$).
\emph{Intuitively, the accuracy drop of caching also implies that we could not use a ``static'' down-sampled graph throughout the training process.}
\textbf{\ding{187}} 
\textit{Surprisingly, jointly applying the caching and switching, the performance drop can be minimized.}



\section{Acknowledgements}
The authors thank the anonymous reviewers for their helpful comments. The work is in part supported by NSF grants NSF IIS-2224843. 
The views and
conclusions contained in this paper are those of the authors and
should not be interpreted as representing any funding agencies.

\vspace{-.5em}
\section{Conclusions and Future work}
We propose \rsc, which replaces the sparse computations in GNNs with their fast approximated versions.
\rsc can be plugged into most of the existing training frameworks to improve their efficiency.
Future work includes exploring \rsc for GNNs that rely on scatter-and-gather operations.

\nocite{langley00}

\bibliography{example_paper}
\bibliographystyle{icml2023}

\newpage
\appendix
\onecolumn
\section{Mathematical Analysis}
\label{app: proof}

\subsection{Why Sampling-based Approximation for GNN?}
\label{app: motivation}

In the main text, we mentioned \sdmm is the main speed bottleneck for GNNs.
Below we illustrate why the column-row sampling is suitable for accelerating \sdmm in GNNs, from the approximation error perspective.
Here we analyze $\Tilde{\mA} \mJ^{(l)}=\sdmm(\Tilde{\mA}, \mJ^{(l)})$ for illustration convenience.
For the backward pass of \sdmm, the analysis is similar, except that we are approximating $\nabla\mJ^{(l)}=\sdmm(\Tilde{\mA}^\top, \nabla\mH^{(l+1)})$.

Column-row sampling approximates the matrix production by excluding some ``unimportant'' columns and rows in the original matrix.
So intuitively, the approximation error $\E[||\Tilde{\mA}\mJ^{(l)}-\texttt{approx}(\Tilde{\mA}\mJ^{(l)})||_F]$ is low if the ``unimportant'' columns/rows are correlated in the selected one. 
Namely, $\Tilde{\mA}$ and $\mJ^{(l)}$ are low-rank.
Formally, we have the following theorem:

\begin{restatable}[\cite{martinsson2020randomized}]{theorem}{error_bound}
\label{theo:error_bound}
Suppose we approximate $\Tilde{\mA}\mJ^{(l)}$ using column-row sampling, and $p_i$ is obtained by \Eqref{eq: col_row_norm}.
Then for any positive number $\epsilon$,
if the number of samples $k$ satisfies
$k \geq \epsilon^{-2}(\srank(\Tilde{\mA})+\srank(\mJ^{(l)}))\log(\mathcal{|V|}+d)$, we have $\E[||\Tilde{\mA}\mJ^{(l)}-\texttt{approx}(\Tilde{\mA}\mJ^{(l)})||_F]\leq 2\epsilon$,
\end{restatable}

where \srank in Theorem \ref{theo:error_bound} is called the stable rank, which is the continuous surrogate measure for the rank that is largely unaffected by tiny singular values.
Formally for any matrix $\mY$, $\srank(\mY)=\frac{||\mY||_F^2}{||\mY||_2}\leq \texttt{rank}(\mY)$.

Fortunately, most real-world graphs are cluster-structured, which means the adjacency matrix $\Tilde{\mA}$ is low-rank \cite{qiu2021lightne, savas2011clustered}.
The low-rank property of real-world graphs is also wildly reported in previous work \cite{jin2020graph, qiu2021lightne}.
Moreover, the intermediate activations $\mJ^{(l)}$ and the activation gradients are also low-rank, due to the aggregation.
Namely, low-rank means ``correlation'' in the row/column space. 
The embedding (i.e., rows in the activation matrix) of connected nodes tend to close due to the graph propagation, which resulting in the low-rank property of the activation matrix. 
Thus for GNNs, the approximation error is low with a relatively small number of sample $k$.
This perspective is also experimentally verified in the experiment section.

\subsection{Proof of Proposition 1}
\propunbias*

Here we note that in the main text, for the notation convenience, we ignore the backward pass of ReLU. 
However, the proof here will consider the non-linear activation function to prove the unbiasedness.
Let $\mH^{(l+1)}_{pre}=\sdmm(\Tilde{\mA}, \mJ^{(l)})$ be the pre-activation.
The backward pass of \relu is:

\begin{align}
\label{eq: approx_bwd}
\E[\nabla \mH^{(l+1)}_{pre}] 
&=\E[\mathbbm{1}_{\mH^{(l+1)}_{pre}>0}\odot\nabla \mH^{(l+1)}]\nonumber\\
&=\mathbbm{1}_{\mH^{(l+1)}_{pre}>0}\odot\E[\nabla \mH^{(l+1)}],
\end{align}

where $\odot$ is the element-wise product and $\mathbbm{1}$ is the indicator function.
The element-wise product is linear operation and $\mathbbm{1}_{\mH^{(l+1)}_{pre}>0}$ is only related to the pre-activation in the forward pass, we only apply the approximation during the backward pass
so $\mathbbm{1}_{\mH^{(l+1)}_{pre}>0}$ can be extracted from the expectation.
We know that for the last layer, we have $\E[\nabla \mH^{(L)}]=\mH^{(L)}$ since we do not apply ReLU at the output layer.
We then can prove by induction that $\E[\nabla \mH^{(l+1)}]=\mH^{(l+1)}$ and $\E[\nabla\mJ^{(l)}]=\E[\texttt{approx}(\Tilde{\mA}^\top\nabla\mH^{(l+1)}_{pre})]=\nabla\mJ^{(l)}$ for any layer $l$.

\subsection{Analysis of MEAN aggregator}
\label{app: sage_analysis}
For GraphSAGE, one commonly used aggregator is the MEAN aggregator, which can be expressed as follows:

\begin{equation}
    \mH^{(l+1)} = \mW_1 \mH^{(l)} + \mW_2 \sdmmean(\mA, \mH^{(l)}),
\end{equation}

where \sdmmean is one variant of the vanilla \sdmm, which replace the reducer function from \texttt{sum($\cdot$)} to \texttt{mean($\cdot$)}. 
We note that in popular GNN packages, the MEAN aggregator usually is implemented based on \sdmmean \cite{pyg, dgl} to reduce the memory usage.
Here we give an example of \sdmmean to illustrate how it works:

\begin{equation}
\sdmmean(
\begin{bmatrix}
1 & 0 \\
0 & 4 \\
5 & 6
\end{bmatrix}, 
\begin{bmatrix}
7 & 8 \\
9 & 10 \\
\end{bmatrix}) = 
\small
\begin{bmatrix}
\frac{1}{2}(1\times7+0\times9) &  \frac{1}{2}(1\times8+0\times10)\\
\frac{1}{2}(0\times7+4\times9) & \frac{1}{2}(0\times8+4\times10) \\
\frac{1}{2}(5\times7+6\times9) & \frac{1}{2}(5\times8+6\times10) 
\end{bmatrix}\nonumber,
\end{equation}

Equivalently, the \sdmmean can also be expressed as:
\begin{equation}
    \sdmmean(\mA, \mH^{(l)}) = \mD^{-1}\mA\mH^{(l)}\nonumber,
\end{equation}

where $\mD$ is the degree matrix of $\mA$.
Thus, although we did not normalize the adjacency matrix in GraphSAGE, when applying the top-$k$ sampling to approximate $\sdmmean$, 
the column norm of $\mA_{:, j_i}$ is actually $\frac{1}{\sqrt{\mathrm{Deg}_{j_i}}}$ due to the normalization.

Also, for GraphSAGE, the inputs to the first \sdmmean operation are $\mA$ and $\mX$.
They do not require gradient since they are not trainable.
Thus, the first SAGE layer is not presented in Figure \ref{fig: ablation_node_deg} and Figure \ref{fig: ablation_layer-wise_ratio}.

\section{Pseudo code of the greedy algorithm}
\label{app: pseudo_code}

\begin{algorithm*}[ht!]
\caption{The greedy algorithm}
\label{algo:greedy}
\begin{algorithmic}
\STATE {\bfseries Inputs:} Gradients of node embeddings $\{\nabla \mH^{(1)}, \cdots \nabla\mH^{(L)}\}$, adjacency matrix $\mA$, graph $\gG=(\gV, \gE)$, hidden dimensions $\{d_1,\cdots d_L\}$.

\STATE {\bfseries Parameters:} The step size $\alpha$, the overall budget $C$.

\STATE {\bfseries Outputs:} The layer-wise $\{k_1,\cdots k_L\}$ associated with the top-$k$ sampling.

\STATE $B \leftarrow \sum_{l=1}^{L}|\mathcal{E}| d_l$.

\STATE $\forall i, k_l \leftarrow |\mathcal{V}|, \mathrm{Top}_{k_l}\leftarrow\{1,\cdots |\mathcal{V}|\}$.

\WHILE{$B\geq C\sum_{l=1}^{L}|\mathcal{E}| d_l$}
\STATE $m \leftarrow \argmin_{l\in\{1,\cdots L\}} ( 
\sum_{i \in \mathrm{Top}_{k_l}}\frac{\|\mA_{:,i}^\top\|_2 \|\nabla\mH^{(l+1)}_{i,:}\|_2}{\|\mA\|_F\|\nabla\mH^{(l+1)}\|_F} - \sum_{i \in \mathrm{Top}_{k_l-\alpha|\mathcal{V}|}}\frac{\|\mA_{:,i}^\top\|_2 \|\nabla\mH^{(l+1)}_{i,:}\|_2)}{\|\mA\|_F\|\nabla\mH^{(l+1)}\|_F}$ 
\hfill \texttt{/*} Choose the layer $m$ to reduce by a step size $\alpha|\mathcal{V}|$, such that the increment of errors is minimal. \texttt{*/}
\STATE $B\leftarrow B - d_m\sum_{i\in\mathrm{Top}_{k_m}\cap i\notin\mathrm{Top}_{k_m - \alpha|\mathcal{V}|}} \#nnz_{i}$ 
\hfill \texttt{/*}Since we exclude some column-row pairs for layer $m$, here we reduce the budget $B$ accordingly. \texttt{*/}
\STATE $k_m\leftarrow k_m - \alpha |\mathcal{V}|$ \hfill \texttt{/*} Update $k_m$ accordingly. \texttt{*/}
\STATE $\mathrm{Top}_{k_m}\leftarrow$ the set of indices $i$ associated with $k_m$ largest $\frac{\|\mA_{:,i}^\top\|_2 \|\nabla\mH^{(l+1)}_{i,:}\|_2}{\|\mA\|_F\|\nabla\mH^{(l+1)}\|_F}$ \hfill \texttt{/*} Update $\mathrm{Top}_{k_m}$ accordingly. \texttt{*/}
\ENDWHILE

\STATE {\bfseries Return} {$\{k_1,\cdots, k_L\}$}

\end{algorithmic}
\end{algorithm*}

In algorithm \ref{algo:greedy}, here we provide the pseudo code of our greedy algorithm for solving the constrained optimization problem.
In Table \ref{tab: greedy_runtime}, we show the run time of the greedy algorithm, which is negligible compared to the acceleration effect.

\section{Extended Related works}
\label{app: related_work}

\noindent
\textbf{Connections to Graph Data Augmentation}
\label{sec: connect_graphda}
Data augmentation \cite{liu2021divaug, han2022g} is wildly adopted in the graph learning for improving model generalization, including dropping nodes \cite{grand}, dropping edges \cite{dropedge}, and graph mixup \cite{han2022g}.
As shown in Figure \ref{fig:csr_example}, the top-$k$ sampling drops the entire columns in the adjacency matrix, while keeping the number of rows unchanged.
\textbf{
That means \rsc drops all of the out edges for a set of nodes.}
This can be viewed as the ``structural dropedge'' for improving the efficiency.
Since we only apply the top-$k$ sampling in the backward pass and top-$k$ indices are different for each operation,
\textbf{
\rsc essentially forward pass with the whole graph, backward pass with different subgraphs at each layer.}
This structural dropedge and heterogeneous backward propagation introduce the regularization effect.
Thus as shown in the experiment section, \rsc may also improve the model accuracy over the baseline.

\noindent
\textbf{Subgraph-based GNN training.}
The key idea of this line of work is to improve the scalability of GNNs by separating the graph into overlapped small batches, then training models with sampled subgraphs \cite{gsage, huang2018adaptive, ladies, clustergcn, gsaint}.
Based on this idea, various sampling techniques have been proposed, including the node-wise sampling \cite{gsage, s-gcn}, layer-wise sampling \cite{huang2018adaptive, ladies}, and subgraph sampling \cite{clustergcn, gsaint}.
However, this approach reduces the memory footprint but results in extra time cost to compute the overlapping nodes between batches.
Generally, methods in this category are orthogonal to \rsc, and they can be combined.

\noindent
\textbf{Graph precomputation.}
The graph precomputation methods decouple the message passing from the model training,
either as a preprocessing step \citep{sgc, appnp, sign} or post-processing step \citep{correct_smooth},
 where the model is simplified as the Multi-Layer Perceptron (MLP).
We did consider this line of work in this paper since the backbone model is not GNN anymore.

\noindent
\textbf{Distributed GNN training.}
The distributed training leverages extra hardwares to increase the memory capacity and training efficiency \cite{zha2023mlsys, dreamshard, YuanWDTKJ22, wang2022dragonn, wangcupcake}. However, the graph data cannot be trivially divided into independent partitions due to the node connectivity. 
Thus, the graph distributed training frameworks propose to split graph into related partitions and minimize the communication overhead \cite{bdsgcn, wan2022pipegcn, ramezani2022learn}. 
Our methods are orthogonal to this line of work.

\noindent
\textbf{Other randomized GNN training.}
Dropedge \cite{dropedge} randomly drops edges to avoid the over-smoothing problem.
Graph Random Neural Networks (Grand) \cite{grand} randomly drop nodes to generate data augmentation for improving model generalization.
Grand+ improves the scalability over Grand by \textbf{pre-computing} a general propagation matrix
and employ it to perform data augmentation \cite{grand+}.
As shown in Section \ref{sec: connect_graphda},
the key difference between GRAND(+) and \rsc is that \emph{\rsc does not drop any node.
Instead \rsc drops all of the out edges for a set of nodes only during backward pass.
Moreover, the drop pattern are evolving during the training process.}
This can be viewed as the ``structural dropedge''.
However, unlike Dropedge \cite{dropedge}, \rsc drop the column-row pairs according to the euclidean norm instead of uniformly dropping.

\section{Experimental Settings}
\label{app: exp_setting}

\subsection{Software and Hardware Descriptions}
All experiments are conducted on a server with four NVIDIA 3090 GPUs, four AMD EPYC 7282 CPUs, and 252GB host memory.
We implement all models based on Pytorch and Pytorch Geometric.
During our experiments, we found that the version of Pytorch, Pytorch Sparse, and Pytorch Scatter can significantly impact the running speed of the baseline.
Here we list the details of our used packages in all experiments in Table \ref{tab: package config}.

\begin{table}[h!]
\centering
\caption{Package configurations of our experiments.}
\label{tab: package config}
\begin{tabular}{ccc} 
\hline
Package & Version \\
\hline
CUDA & 11.1 \\
pytorch\_sparse &  0.6.12\\
pytorch\_scatter & 2.0.8 \\
pytorch\_geometric & 1.7.2  \\
pytorch &  1.9.0\\
OGB & 1.3.2\\
\hline
\end{tabular}
\end{table}

\subsection{Statistics of benchmark datasets}

The statistics for all used datasets are shown in Table \ref{tab:dataset_stat}.
We follow the standard data splits and all datasets are directly downloaded from Pytorch Geometric or the protocol of OGB \citep{ogb}.

\begin{table}[h!]
\centering
\caption{Dataset Statistics.}
\label{tab:dataset_stat}
\begin{tabular}{cccccc} 
\hline
Dataset                & Task         & Nodes     & Edges      & Classes & Label Rates  \\ 
\hline
Reddit                 & multi-class  & 232,965   & 11,606,919 & 41      & 65.86\%      \\
Yelp                   & multi-label  & 716,847   & 6,977,409  & 100     & 75.00\%      \\
\textit{ogbn-proteins} & binary-Class & 132,534   & 39,561,252 & 2       & 65.00\%      \\
\textit{ogbn-products} & multi-class  & 2,449,029 & 61,859,076 & 47      & 8.03\%       \\
\hline
\end{tabular}
\end{table}

\subsection{Hyperparameter Settings}

Regarding Reddit and Yelp dataset, we follow the hyperparameter reported in the respective papers as closely as possible.
Regarding \textit{ogbn-proteins} and \textit{ogbn-products} dataset, we follow the hyperparameter configurations and codebases provided on the OGB \cite{ogb} leader-board.
Please refer to the OGB website for more details. 
The optimizer is Adam for all these models.
All methods terminate after a fixed number of epochs. 
We report the test accuracy associated with the highest validation score.
Table \ref{tab: graphsaint hp} summarize the hyperparameter configuration of GraphSAINT.
Table \ref{tab: fullbatch gcn}, Table \ref{tab: fullbatch sage}, and Table \ref{tab: fullbatch gcn2} summarize the hyperparameter configuration of full-Batch GCN, GraphSAGE, and GCNII, respectively.

\begin{table}[h!]
\centering
\caption{Configuration of Full-Batch GCN.}
\label{tab: fullbatch gcn}
\begin{tabular}{c|ccc|ccc} 
\hline
\multirow{2}{*}{Dataset}                                                  & \multicolumn{3}{c|}{Training}                                              & \multicolumn{3}{c}{Archtecture}                                                 \\
                                                                          & \begin{tabular}[c]{@{}c@{}}Learning\\Rates\end{tabular} & Epochs & Dropout & BatchNorm & Layers & \begin{tabular}[c]{@{}c@{}}Hidden\\Dimension\end{tabular}  \\ 
\hline
Reddit                                                                    & 0.01                                                    & 400    & 0.5     & Yes       & 3      & 256                                                        \\
Yelp                                                                      & 0.01                                                    & 500    & 0.1     & Yes       & 3      & 512                                                        \\
\begin{tabular}[c]{@{}c@{}}\textit{ogbn-}\\\textit{proteins}\end{tabular} & 0.01                                                    & 1000   & 0.5     & No        & 3      & 256                                                        \\
\begin{tabular}[c]{@{}c@{}}\textit{ogbn-}\\\textit{products}\end{tabular} & 0.001                                                   & 500    & 0.5     & No        & 3~     & 256                                                        \\
\hline
\end{tabular}
\end{table}

\begin{table}[h!]
\centering
\caption{Configuration of Full-Batch GraphSAGE.}
\label{tab: fullbatch sage}
\begin{tabular}{c|ccc|ccc} 
\hline
\multirow{2}{*}{Dataset}                                                  & \multicolumn{3}{c|}{Training}                                              & \multicolumn{3}{c}{Archtecture}                                                 \\
                                                                          & \begin{tabular}[c]{@{}c@{}}Learning\\Rates\end{tabular} & Epochs & Dropout & BatchNorm & Layers & \begin{tabular}[c]{@{}c@{}}Hidden\\Dimension\end{tabular}  \\ 
\hline
Reddit                                                                    & 0.01                                                    & 400    & 0.5     & Yes       & 3      & 256                                                        \\
Yelp                                                                      & 0.01                                                    & 500    & 0.1     & Yes       & 3      & 512                                                        \\
\begin{tabular}[c]{@{}c@{}}\textit{ogbn-}\\\textit{proteins}\end{tabular} & 0.01                                                    & 1000   & 0.5     & No        & 3      & 256                                                        \\
\begin{tabular}[c]{@{}c@{}}\textit{ogbn-}\\\textit{products}\end{tabular} & 0.001                                                   & 500    & 0.5     & No        & 3      & 256                                                        \\
\hline
\end{tabular}
\end{table}

\begin{table}[h!]
\centering
\caption{Configuration of Full-Batch GCNII.}
\label{tab: fullbatch gcn2}
\begin{tabular}{c|ccc|ccc} 
\hline
\multirow{2}{*}{Dataset}                                                  & \multicolumn{3}{c|}{Training}                                              & \multicolumn{3}{c}{Archtecture}                                                 \\
                                                                          & \begin{tabular}[c]{@{}c@{}}Learning\\Rates\end{tabular} & Epochs & Dropout & BatchNorm & Layers & \begin{tabular}[c]{@{}c@{}}Hidden\\Dimension\end{tabular}  \\ 
\hline
Reddit                                                                    & 0.01                                                    & 400    & 0.5     & Yes       & 4      & 256                                                        \\
Yelp                                                                      & 0.01                                                    & 500    & 0.1     & Yes       & 4      & 256                                                        \\
\begin{tabular}[c]{@{}c@{}}\textit{ogbn-}\\\textit{proteins}\end{tabular} & 0.01                                                    & 1000   & 0.5     & No        & 4      & 256                                                        \\
\hline
\end{tabular}
\end{table}

\begin{table}[h!]
\centering
\caption{Training configuration of GraphSAINT.}
\label{tab: graphsaint hp}
\begin{tabular}{c|cc|ccc|ccc} 
\hline
\multirow{2}{*}{Dataset}                                                  & \multicolumn{2}{c|}{\begin{tabular}[c]{@{}c@{}}RandomWalk\\Sampler\end{tabular}} & \multicolumn{3}{c|}{Training}                                              & \multicolumn{3}{c}{Archtecture}                                                 \\
                                                                          & Walk length & Roots                                                              & \begin{tabular}[c]{@{}c@{}}Learning\\Rates\end{tabular} & Epochs & Dropout & BatchNorm & Layers & \begin{tabular}[c]{@{}c@{}}Hidden\\Dimension\end{tabular}  \\ 
\hline
Reddit                                                                    & 4           & 8000                                                               & 0.01                                                    & 40     & 0.1     & Yes       & 3      & 128                                                        \\
Yelp                                                                      & 2           & 8000                                                               & 0.01                                                    & 75     & 0.1     & Yes       & 3      & 512                                                        \\
\begin{tabular}[c]{@{}c@{}}\textit{ogbn-}\\\textit{products}\end{tabular} & 3           & 60000                                                              & 0.01                                                    & 20     & 0.5     & No        & 3      & 256                                                        \\
\hline
\end{tabular}
\end{table}

\section{More experiment results}
\label{app: exp_results}

\begin{figure*}[h!]
    \centering
    \begin{subfigure}[h]{0.33\linewidth}
    \includegraphics[width=\linewidth]{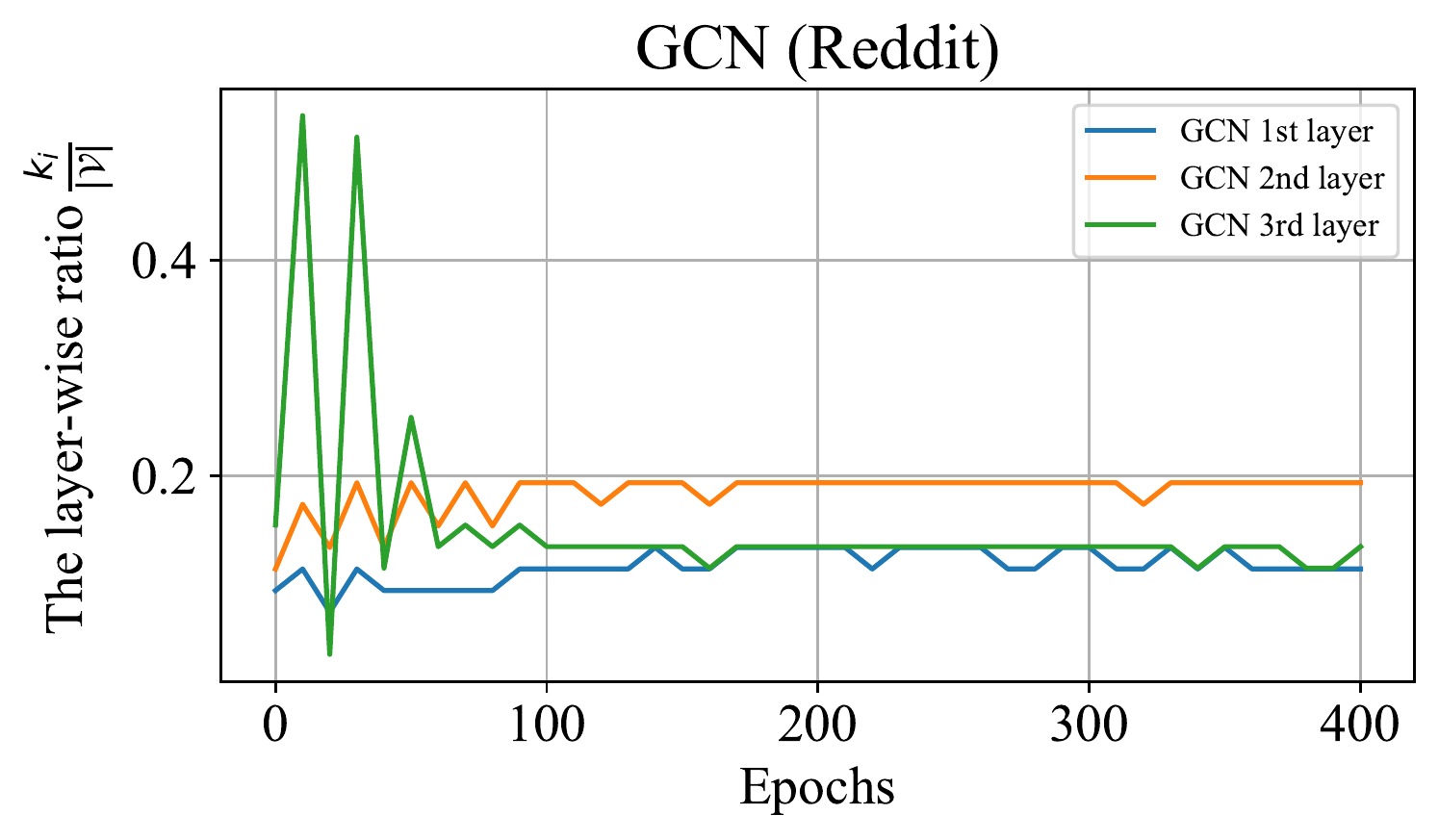}
    \end{subfigure}
    \begin{subfigure}[h]{0.33\linewidth}
    \includegraphics[width=\linewidth]{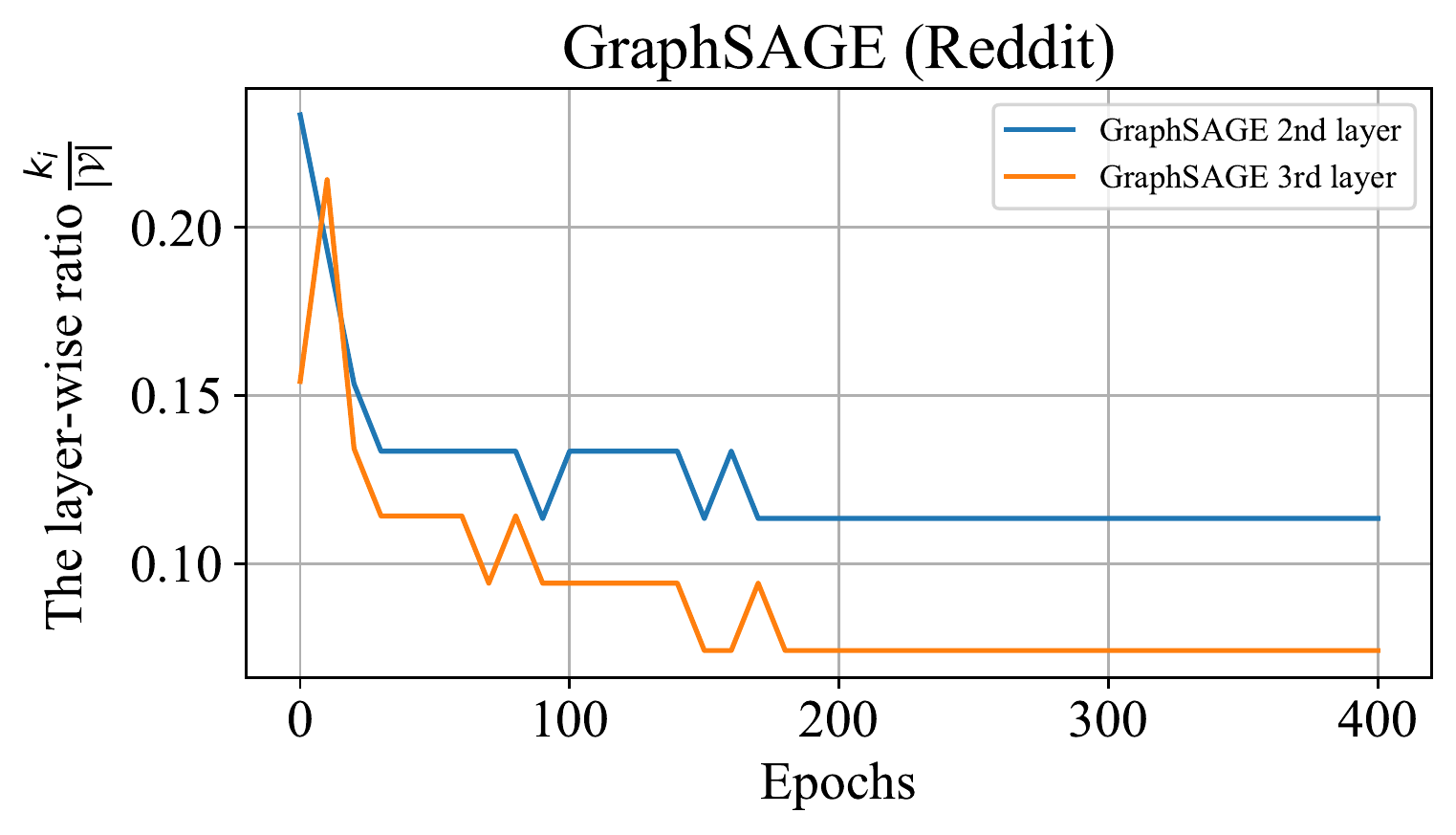}
    \end{subfigure}
    \begin{subfigure}[h]{0.33\linewidth}
    \includegraphics[width=\linewidth]{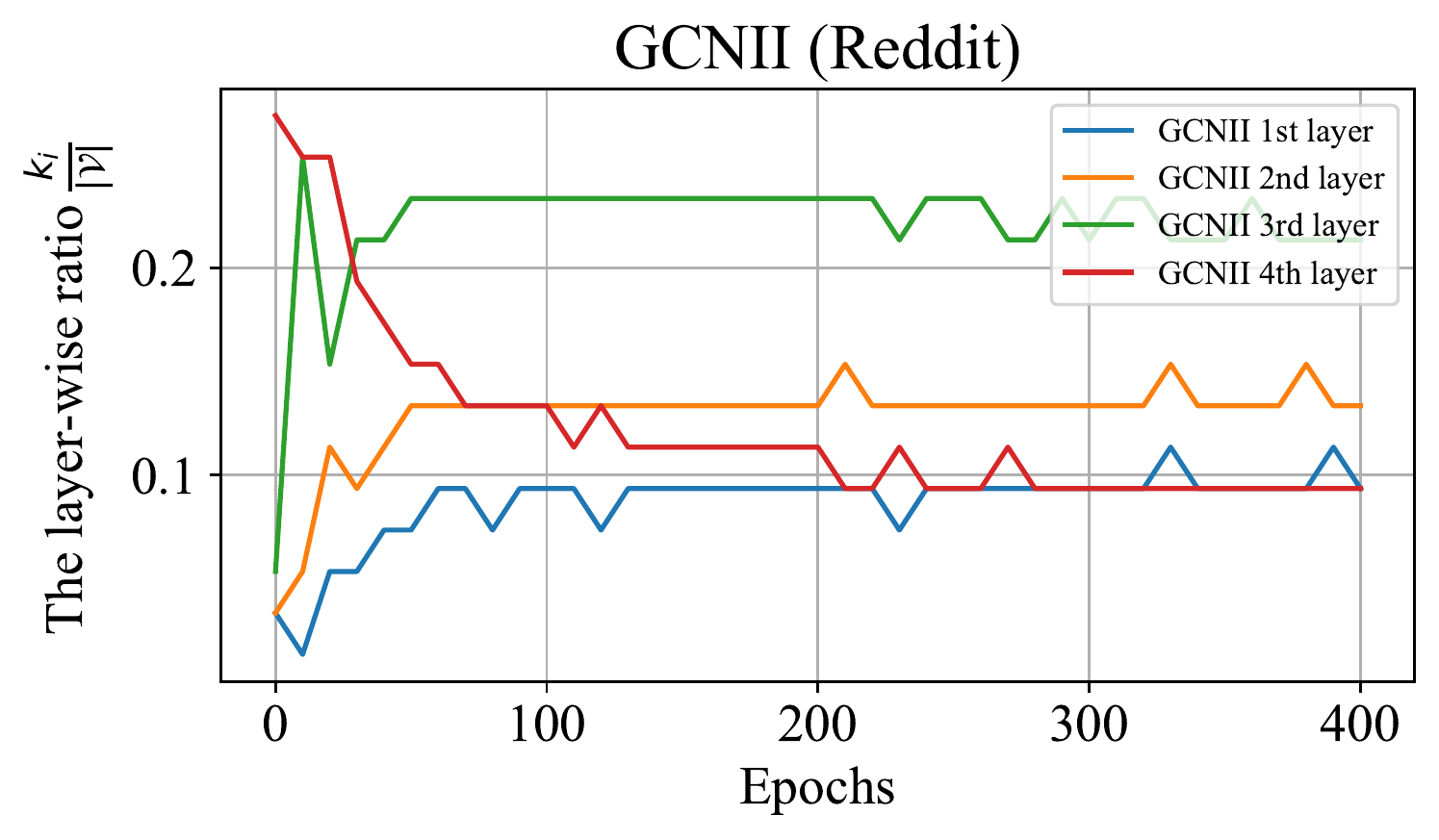}
    \end{subfigure}
     \caption{The allocated layer-wise $k_l$ for GCN, GraphSAGE and GCNII on Reddit, where budget $C$ is set as 0.1.
     The input of the \sdmm in the first GraphSAGE layer does not require gradient and thus absent in the Figure (Appendix \ref{app: sage_analysis}).}
    \label{fig: ablation_layer-wise_ratio}
\end{figure*}

\begin{figure*}[ht!]
    \centering
    \begin{subfigure}[h]{0.33\linewidth}
    \includegraphics[width=\linewidth]{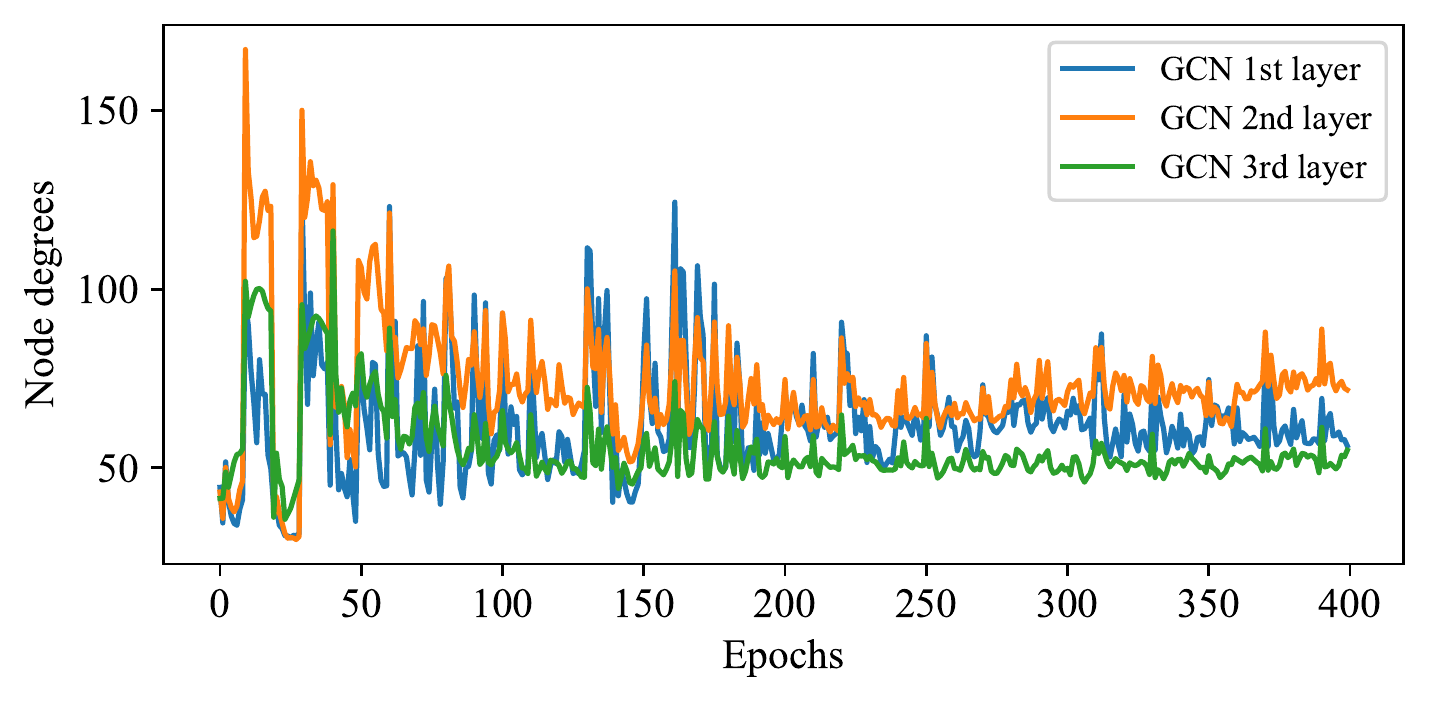}
    \end{subfigure}
    \begin{subfigure}[h]{0.33\linewidth}
    \includegraphics[width=\linewidth]{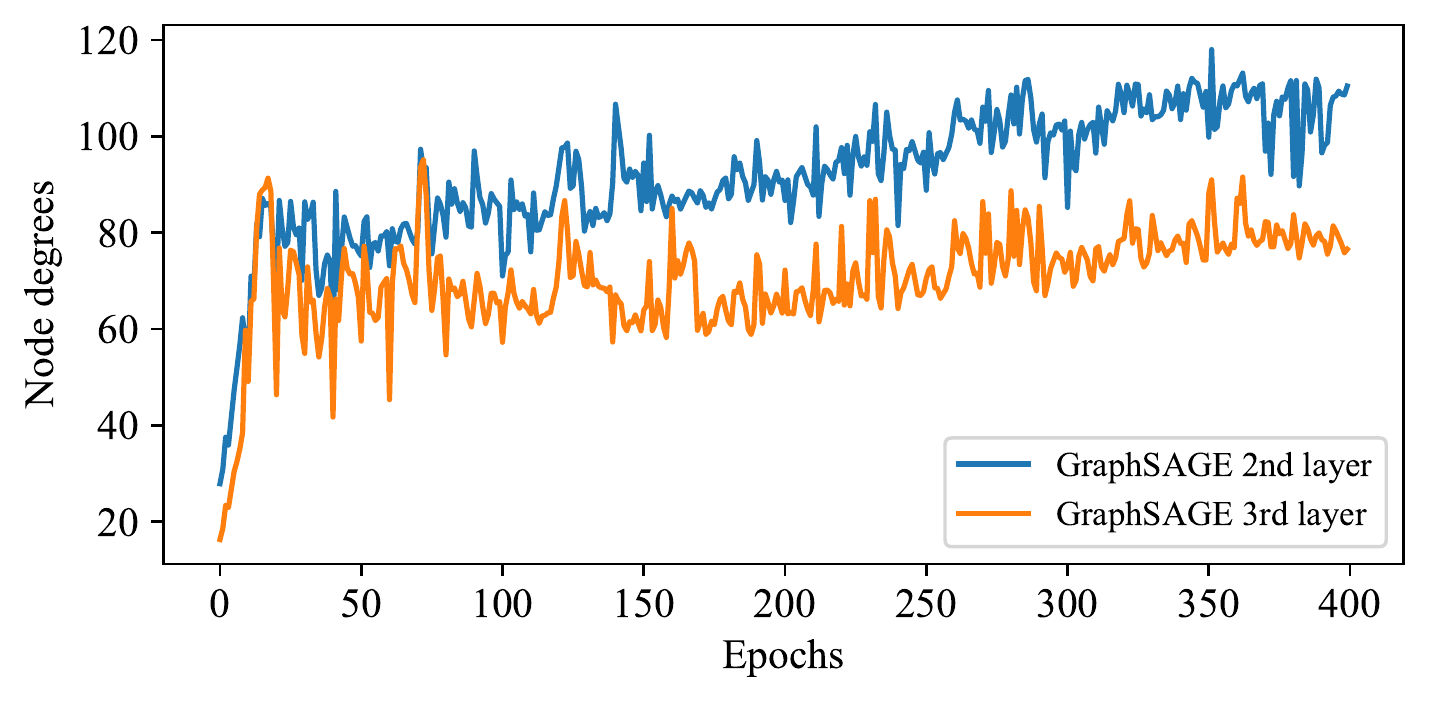}
    \end{subfigure}
    \begin{subfigure}[h]{0.33\linewidth}
    \includegraphics[width=\linewidth]{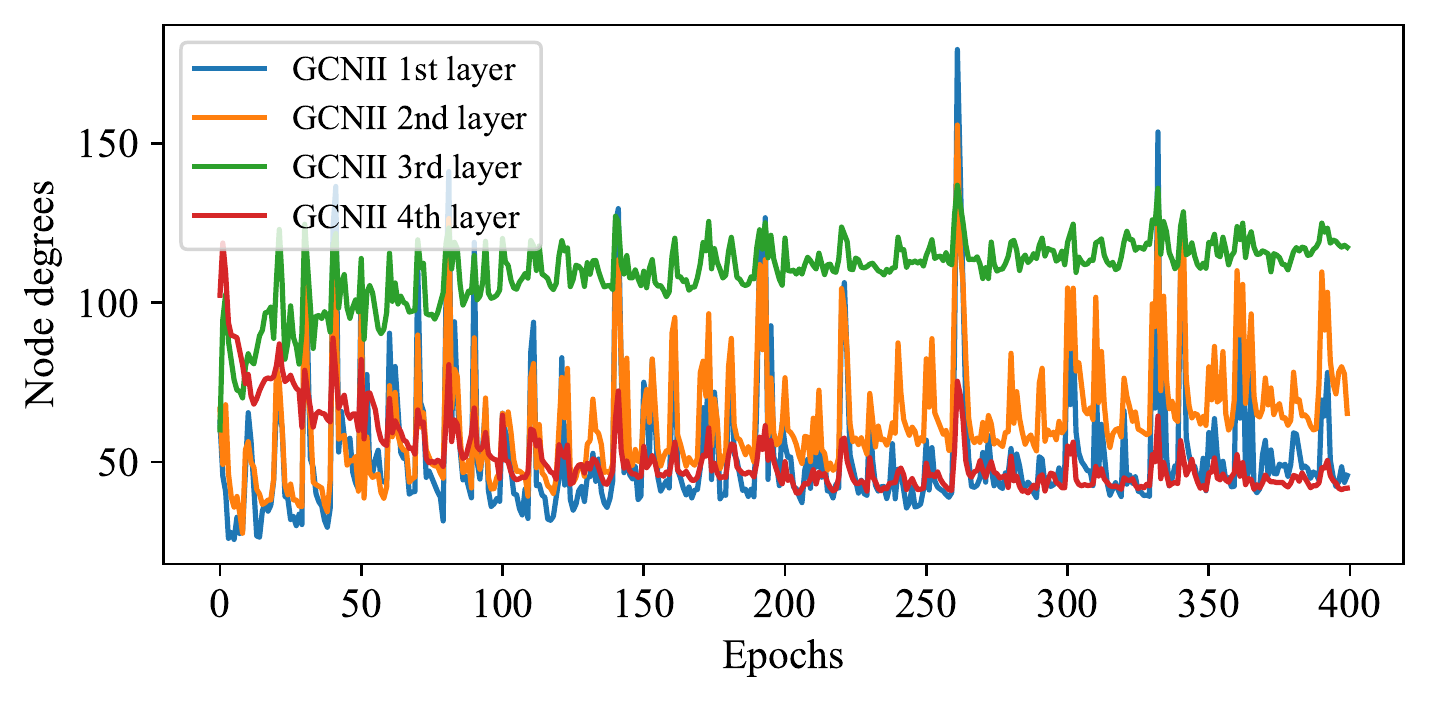}
    \end{subfigure}
     \caption{The averaged degrees of nodes picked by top-$k$ sampling along the whole training process, where the applied dataset is Reddit and overall budget $C$ is set as 0.1.}
    \label{fig: ablation_node_deg}
\end{figure*}

The running time of the greedy algorithm is shown in \ref{tab: greedy_runtime}.
We also visualize the allocated $k_l$ for each layer across iterations in Figure \ref{fig: ablation_layer-wise_ratio}, and the degree of picked nodes in Figure \ref{fig: ablation_node_deg}. Here we use Reddit dataset for the case study.
We observe that the $k_l$ assigned by \rsc evolves along with the training.

\begin{table}[h!]
\centering
\caption{The running time (second) of the greedy algorithm.}
\label{tab: greedy_runtime}
\begin{tabular}{ccccc} 
\hline
          & Reddit & Yelp & \begin{tabular}[c]{@{}c@{}}\textit{ogbn-}\\\textit{proteins}\end{tabular} & \begin{tabular}[c]{@{}c@{}}\textit{ogbn-}\\\textit{products}\end{tabular}  \\ 
\hline
GCN       & 0.03~  & 0.03 & 0.03                                                                      & 0.03                                                                       \\
GraphSAGE & 0.02   & 0.02 & 0.03                                                                      & 0.03                                                                       \\
GCNII     & 0.05   & 0.05 & 0.06                                                                      & -                                                                          \\
\hline
\end{tabular}
\end{table}

\subsection{Additional Ablation Results to the Resource Allocation Algorithm (Figure \ref{fig: ablation_pareto})}
\label{app: more_results_fig_pareto}

\begin{figure*}[h!]
    \centering
    \begin{subfigure}[h]{0.33\linewidth}
    \includegraphics[width=\linewidth]{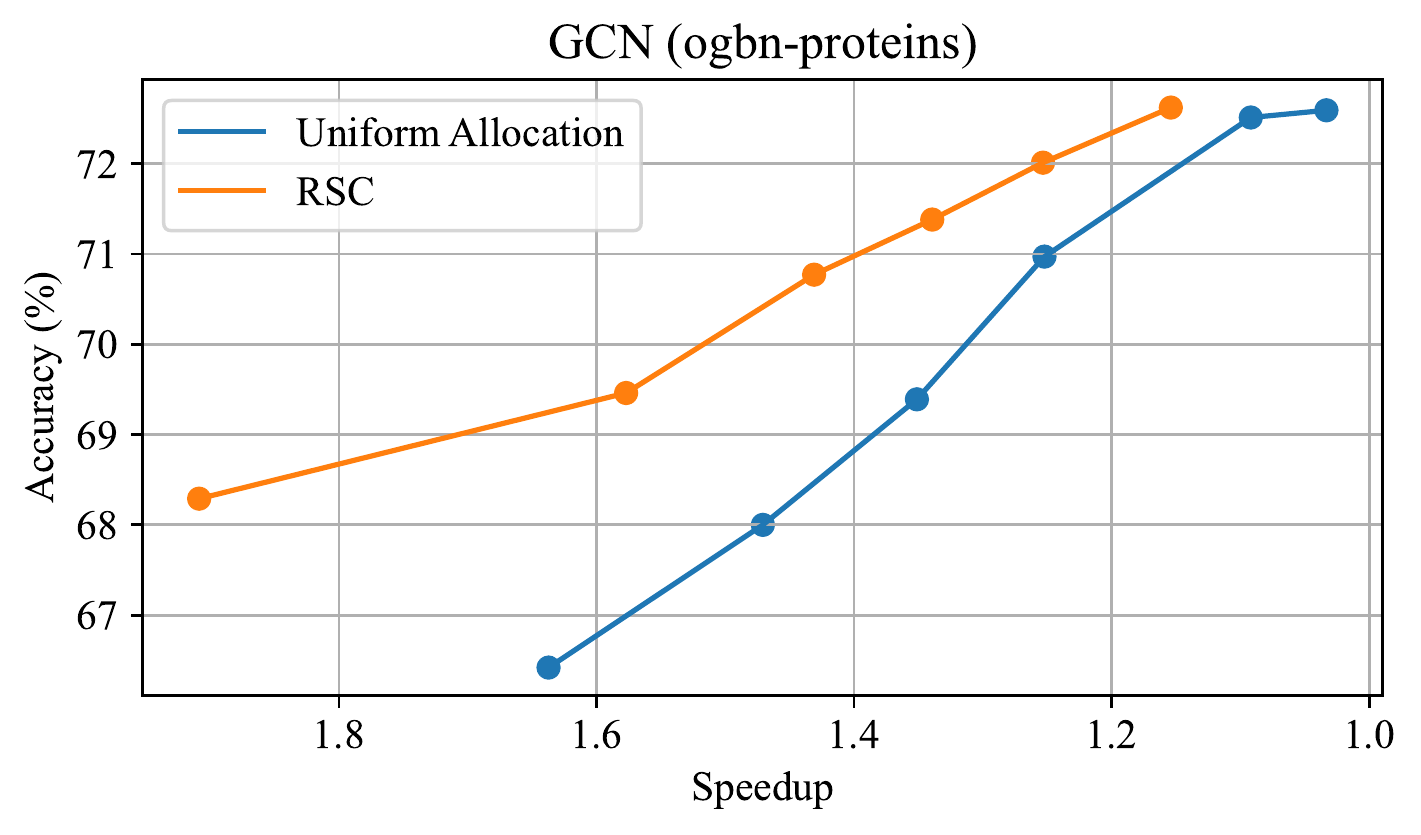}
    \end{subfigure}
    \begin{subfigure}[h]{0.33\linewidth}
    \includegraphics[width=\linewidth]{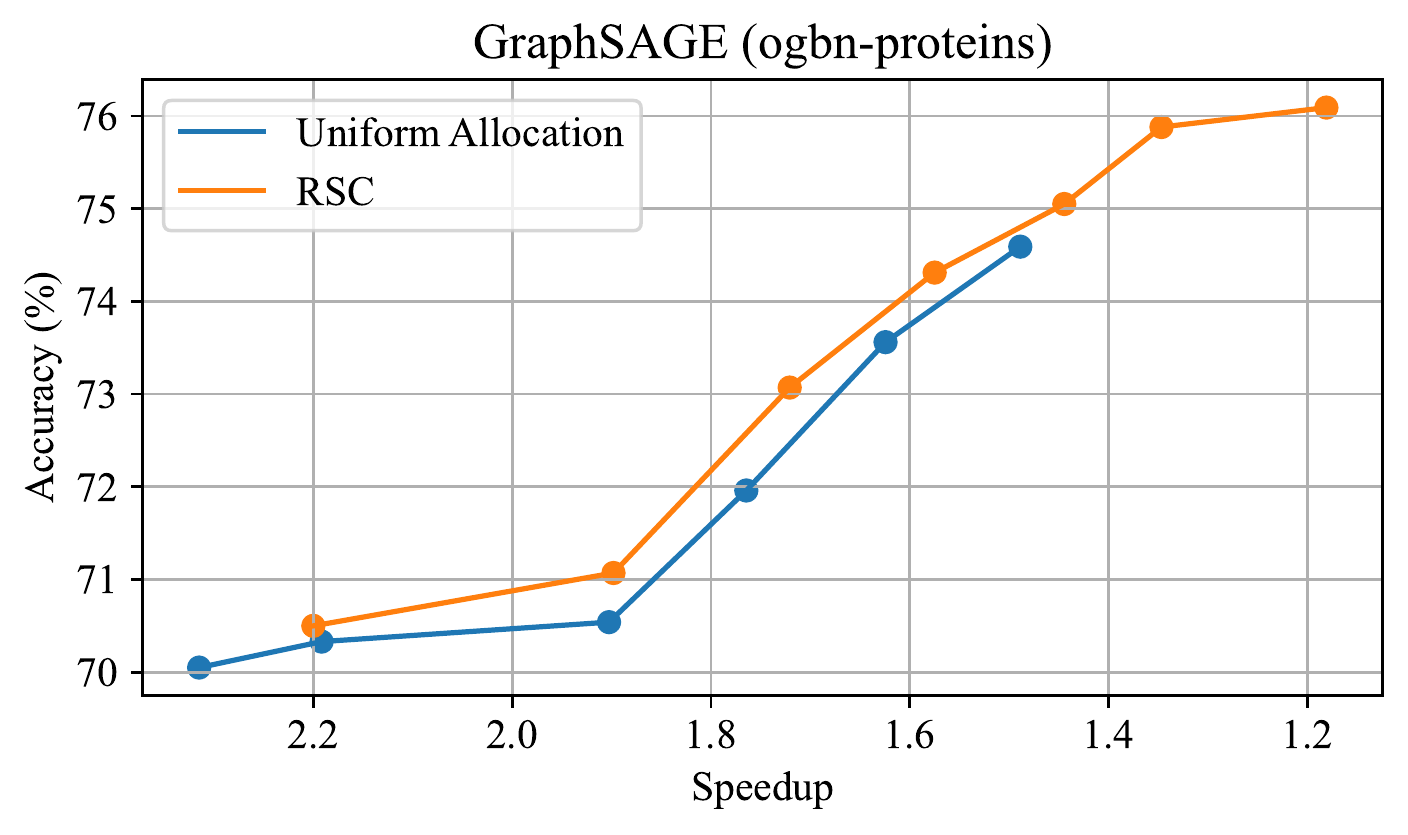}
    \end{subfigure}
    \begin{subfigure}[h]{0.33\linewidth}
    \includegraphics[width=\linewidth]{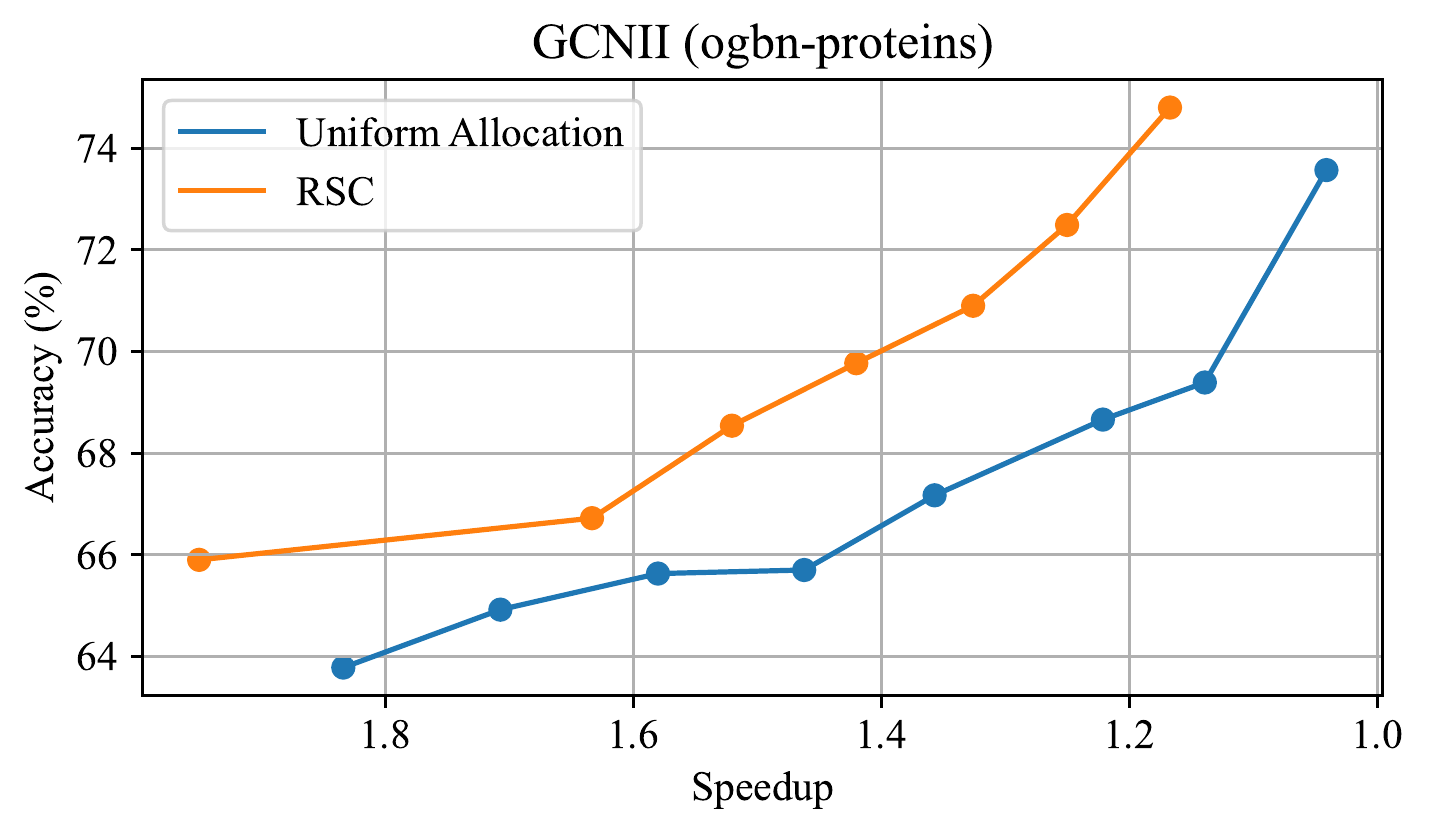}
    \end{subfigure}
    \caption{The Pareto frontier of the accuracy-efficiency trade-off for \rsc and the uniform allocation.
    The dataset is ogbn-proteins.
    Here we disabled the caching and switch mechanism for a fair comparison.}
    \label{fig: ablation_proteins_pareto}
\end{figure*}

\begin{figure*}[h!]
    \centering
    \begin{subfigure}[h]{0.33\linewidth}
    \includegraphics[width=\linewidth]{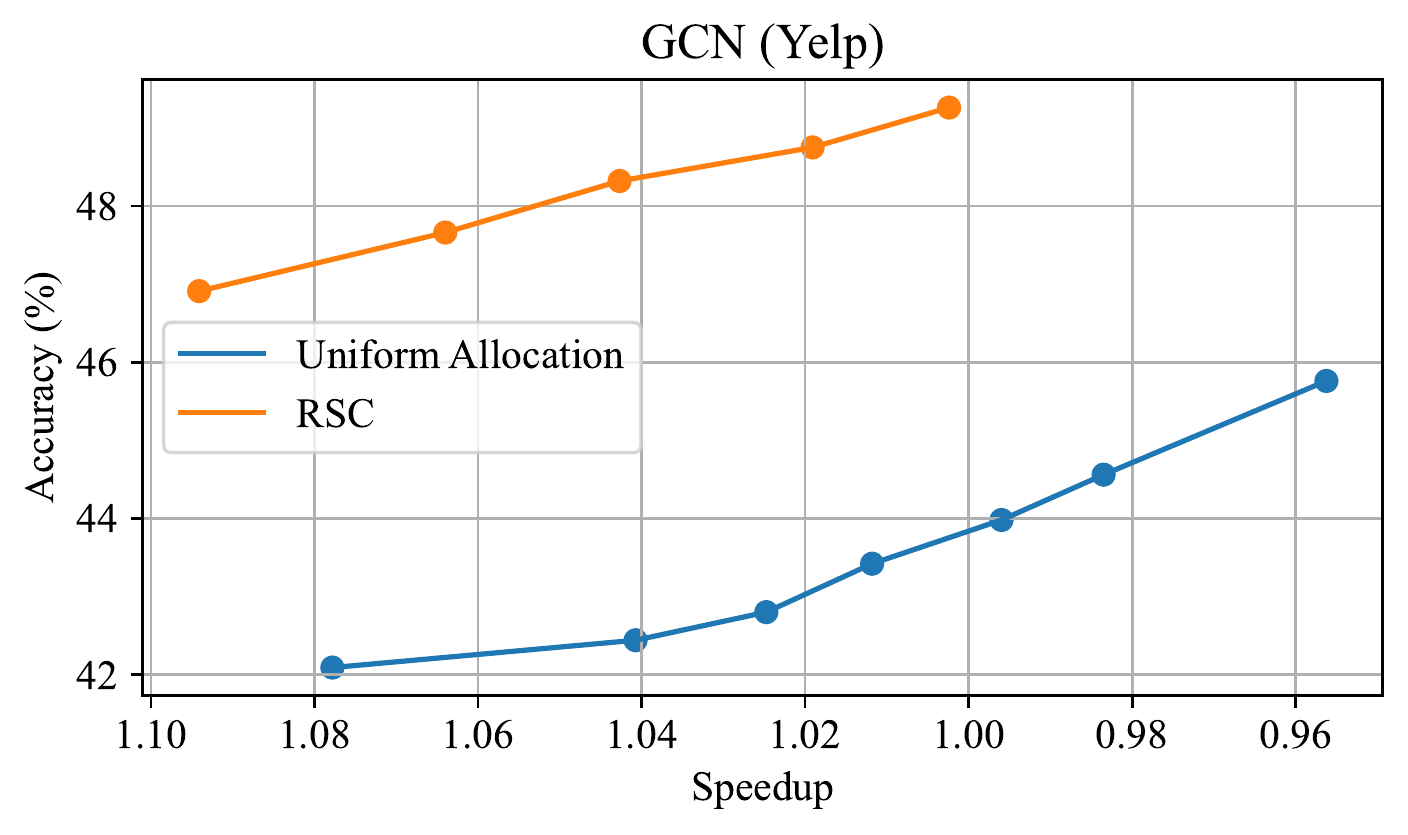}
    \end{subfigure}
    \begin{subfigure}[h]{0.33\linewidth}
    \includegraphics[width=\linewidth]{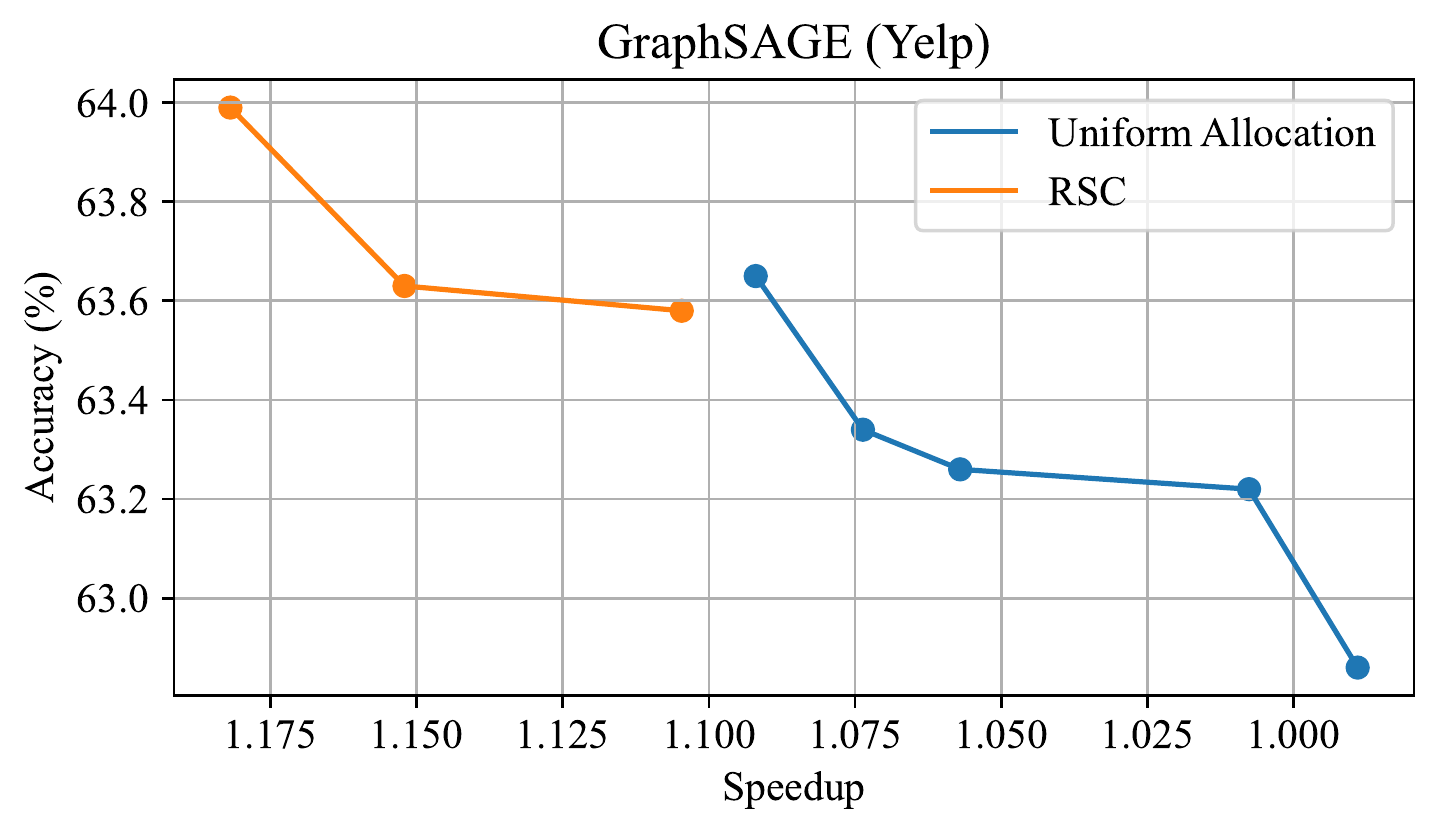}
    \end{subfigure}
    \begin{subfigure}[h]{0.33\linewidth}
    \includegraphics[width=\linewidth]{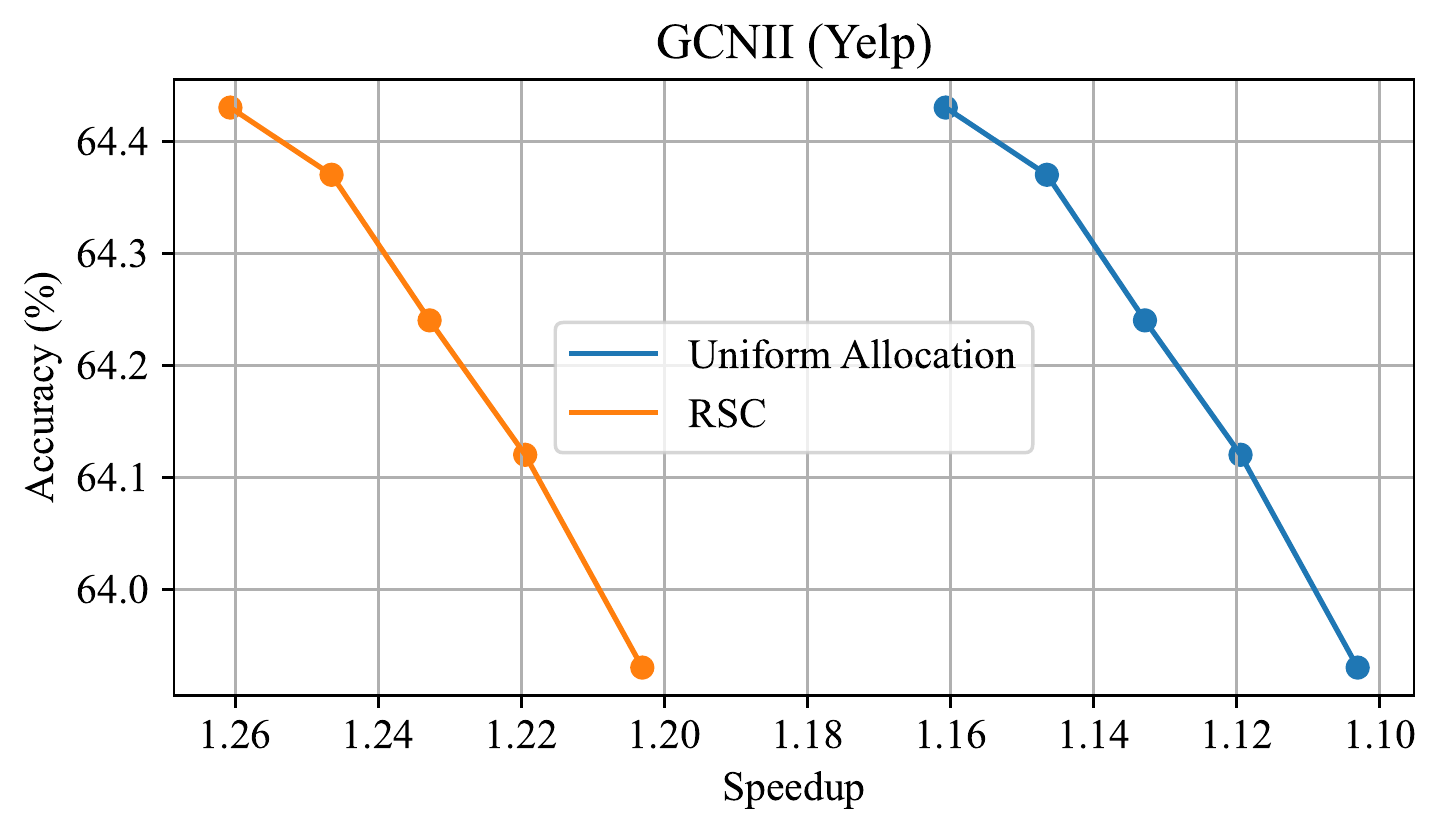}
    \end{subfigure}
    \caption{The Pareto frontier of the accuracy-efficiency trade-off for \rsc and the uniform allocation.
    The dataset is Yelp.
     Here we disabled the caching and switch mechanism for a fair comparison.}
    \label{fig: ablation_pareto_yelp}
\end{figure*}

Due to the page limit, we present more ablation study on the resource allocation algorithm here.
Specifically, in Figure \ref{fig: ablation_proteins_pareto}, we compare \rsc to the uniform allocation on ogbn-proteins dataset with GCN, GraphSAGE, and GCNII, respectively.
In Figure \ref{fig: ablation_pareto_yelp}, we compare \rsc to the uniform allocation on Yelp dataset with GCN, GraphSAGE, and GCNII, respectively.
We conclude that \rsc generally outperforms the uniform allocation strategy.

\subsection{Hyperparameter Sensitivity Analysis}
\begin{figure*}[h!]
    \centering
    \begin{subfigure}[h]{0.33\linewidth}
    \includegraphics[width=\linewidth]{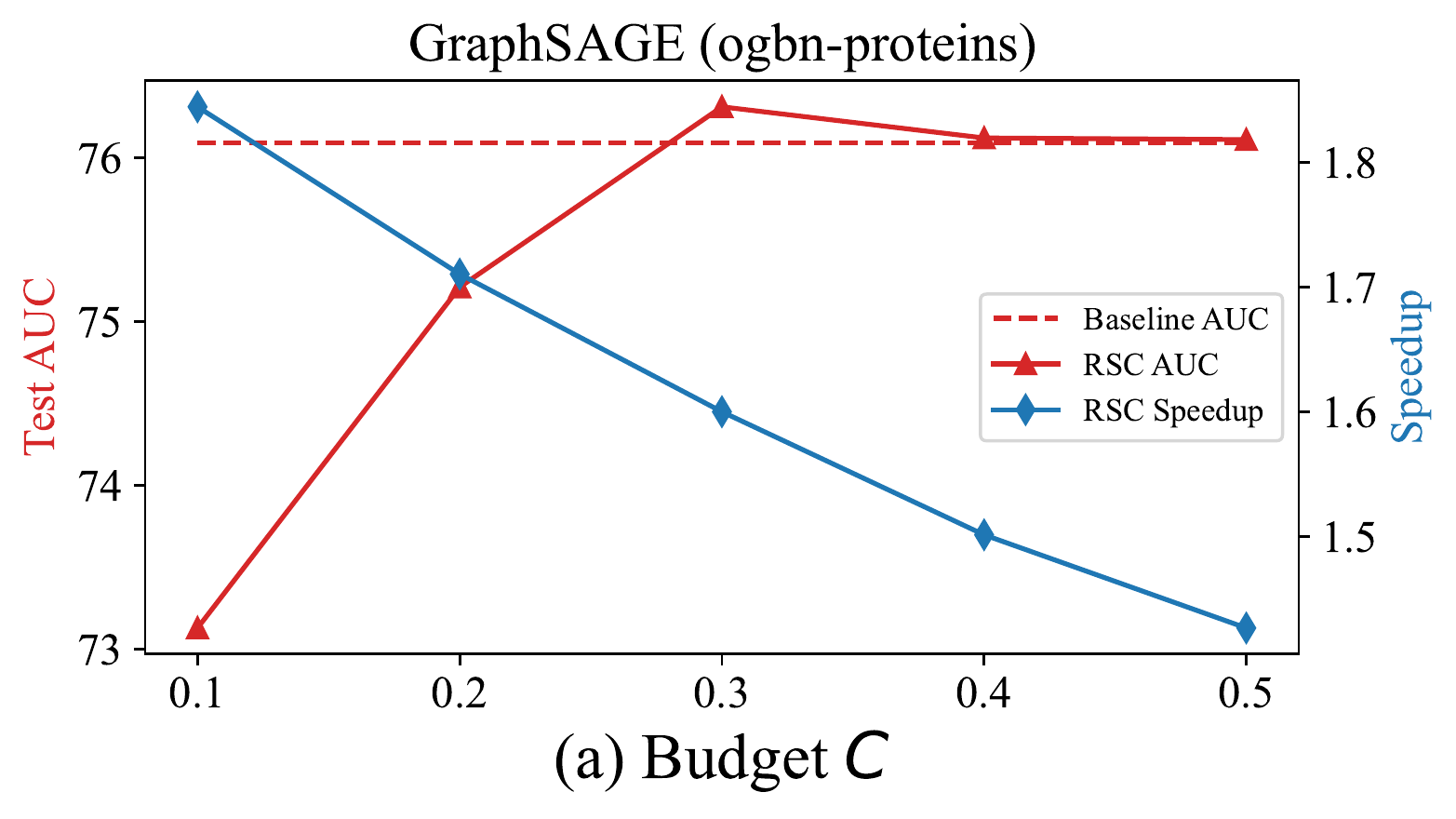}
    \end{subfigure}
    \begin{subfigure}[h]{0.33\linewidth}
    \includegraphics[width=\linewidth]{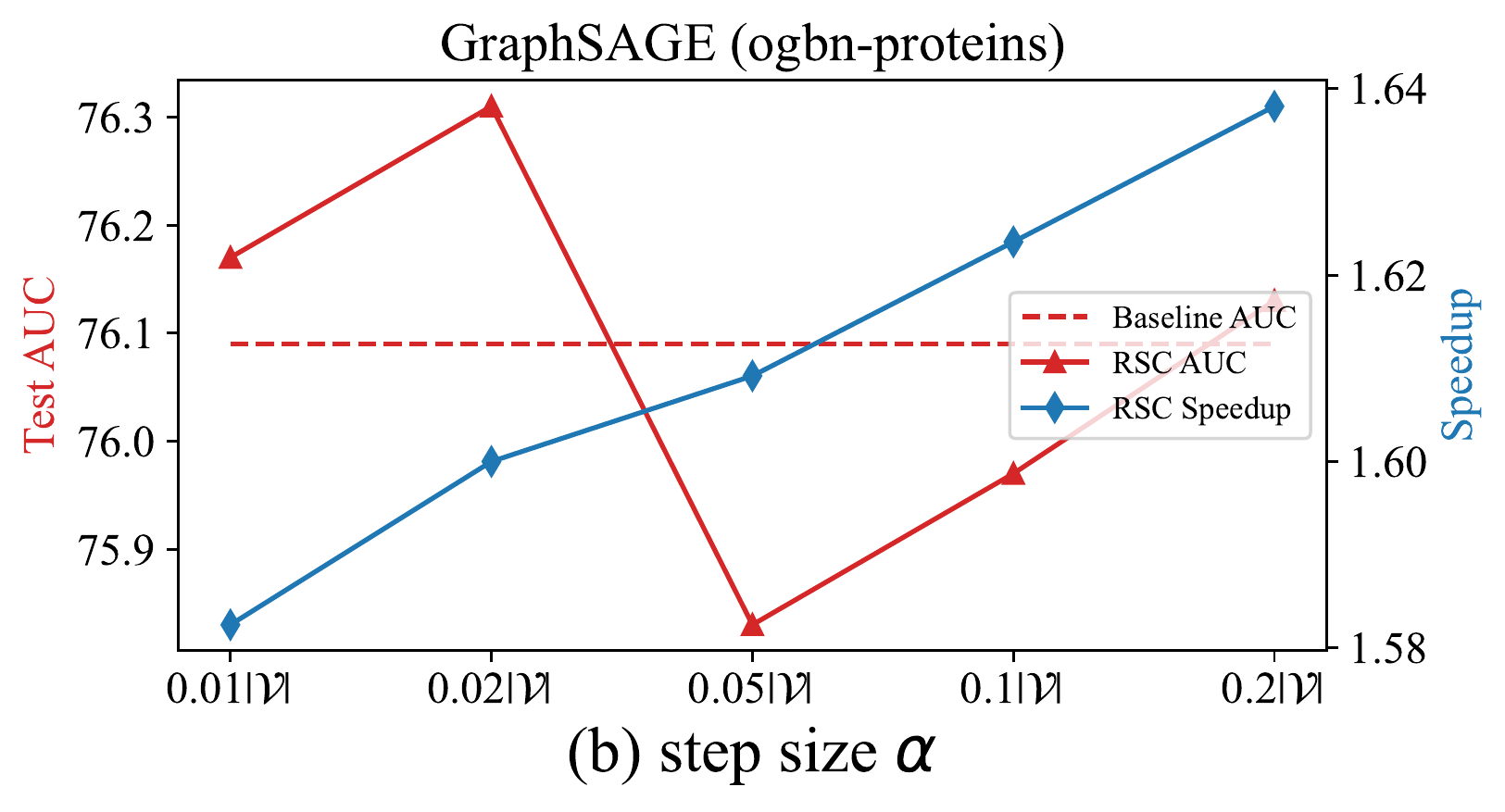}
    \end{subfigure}
    \begin{subfigure}[h]{0.33\linewidth}
    \includegraphics[width=\linewidth]{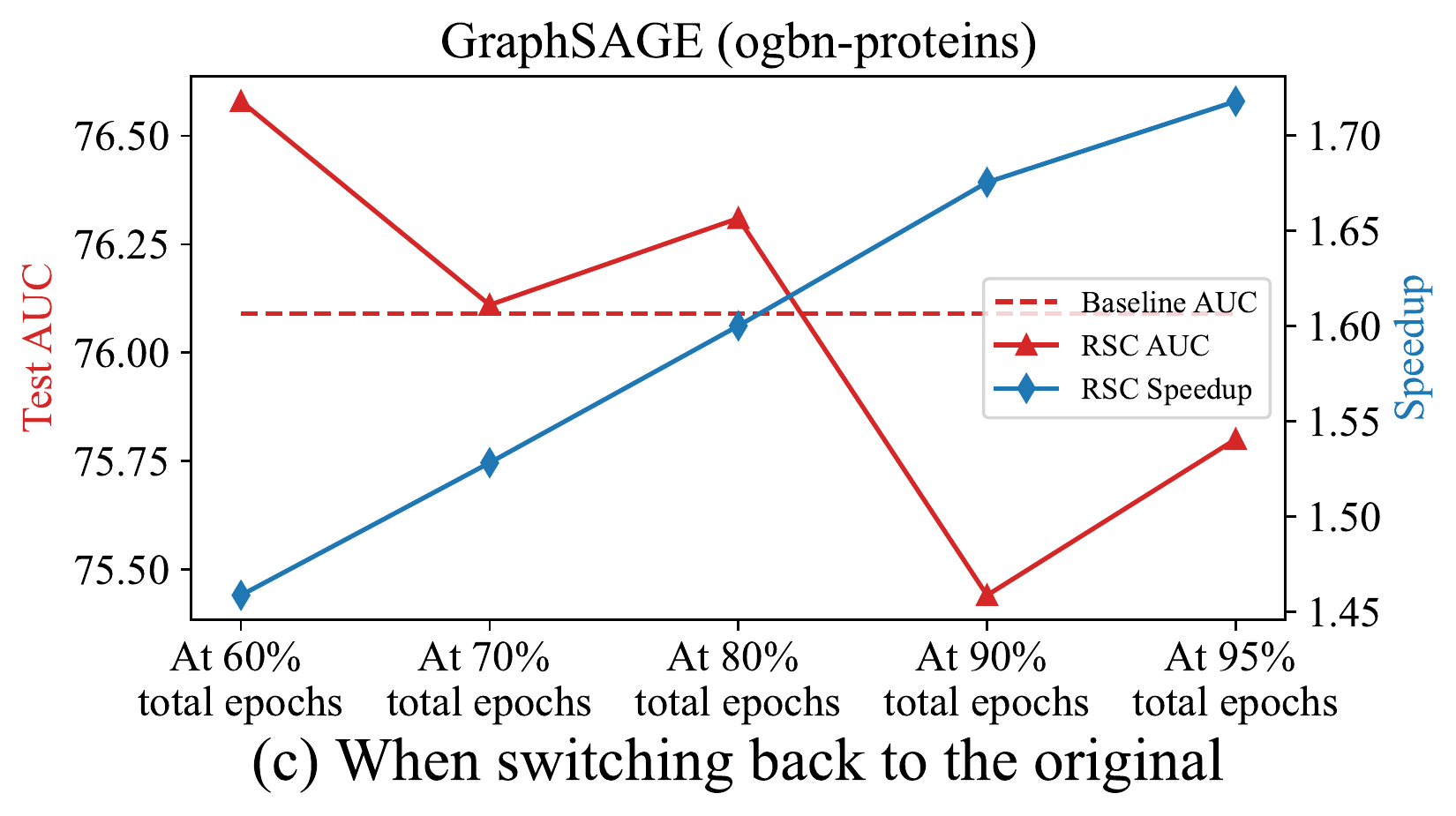}
    \end{subfigure}
    \caption{Hyperparameter analysis w.r.t. the budget $C$, the step size $\alpha$ in Algorithm \ref{algo:greedy}, and when switching back to the original operations. The model is GraphSAGE and the applied dataset is \textit{ogbn-proteins}.}
    \label{fig: hp}
\end{figure*}

Here we analyze the impacts of the main hyperparameters of \rsc:
\textbf{(1)} the budget $C$, which controls the efficiency-accuracy trade-off;
\textbf{(2)} the step size $\alpha$ in the greedy Algorithm \ref{algo:greedy};
\textbf{(3)} when switching back to the original sparse operations.
In Figure \ref{fig: hp}, we vary only one of them with the others fixed.
We conclude
\textbf{(1)} larger budget $C$ leads to better accuracy with smaller speedup, since we are using more computational resources to approximate the full operation.
\textbf{(2)} larger step size $\alpha$ leads to marginally larger speedup since the greedy algorithm will terminate earlier.
Also the step size $\alpha$ does not affect the model accuracy a lot.
In practice, we set $\alpha=0.02|\mathcal{V}|$.
\textbf{(3)} The later we switch back to the original operation, the larger the accuracy drop and the smaller the speedup, it is equivalent to using less resources to approximate the full operation epoch-wisely.
Thus, we apply \rsc for $80\%$ of the total epochs to balance the trade-off.

\begin{figure*}[h!]
    \centering
    \begin{subfigure}[h]{0.33\linewidth}
    \includegraphics[width=\linewidth]{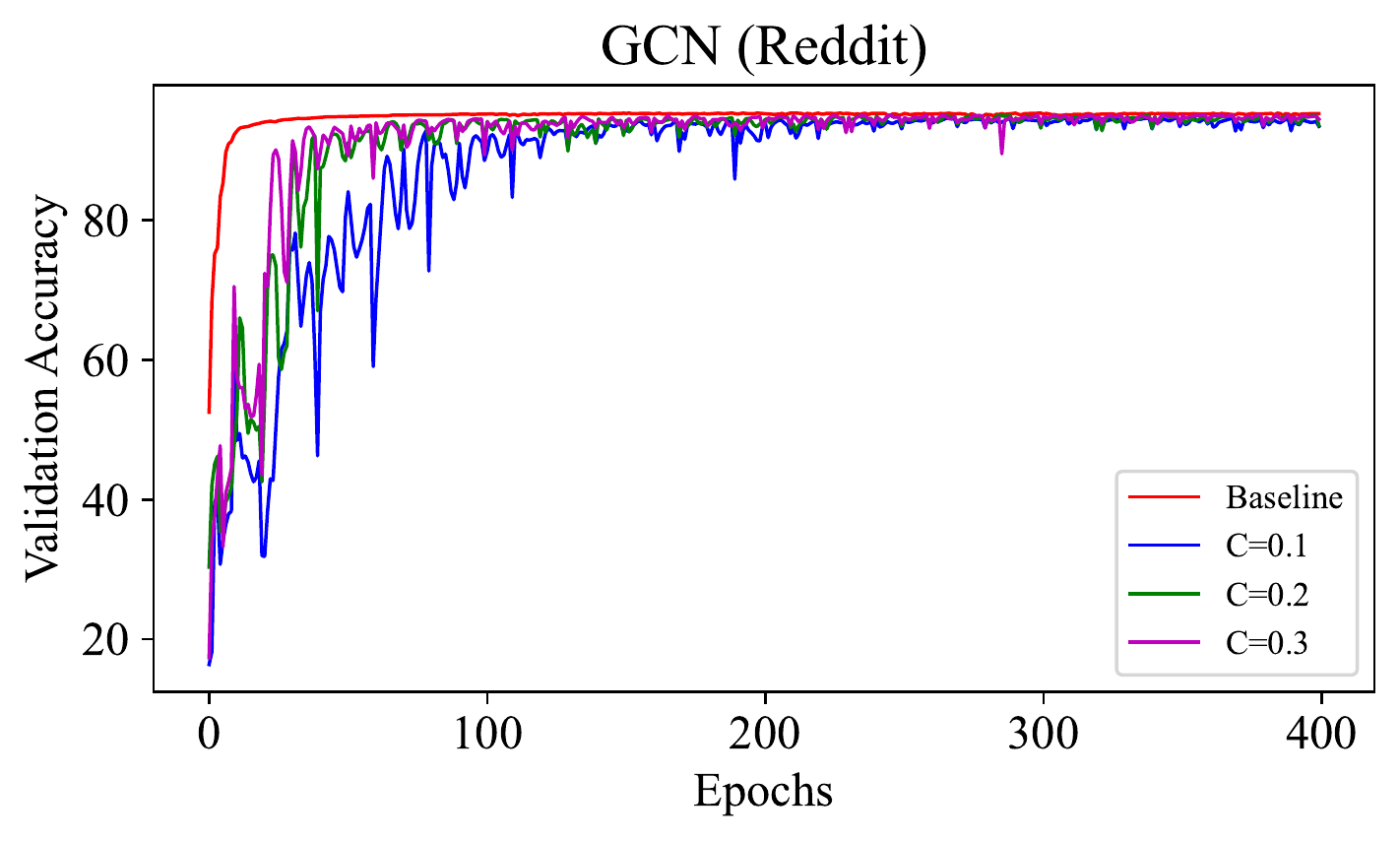}
    \end{subfigure}
    \begin{subfigure}[h]{0.33\linewidth}
    \includegraphics[width=\linewidth]{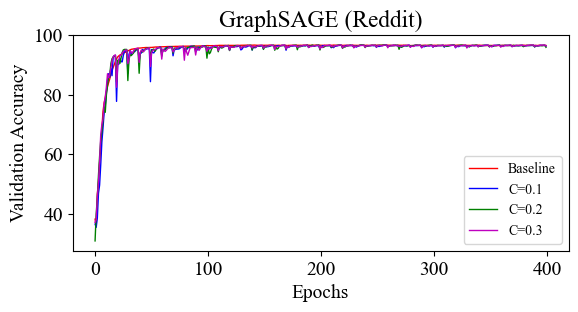}
    \end{subfigure}
    \begin{subfigure}[h]{0.33\linewidth}
    \includegraphics[width=\linewidth]{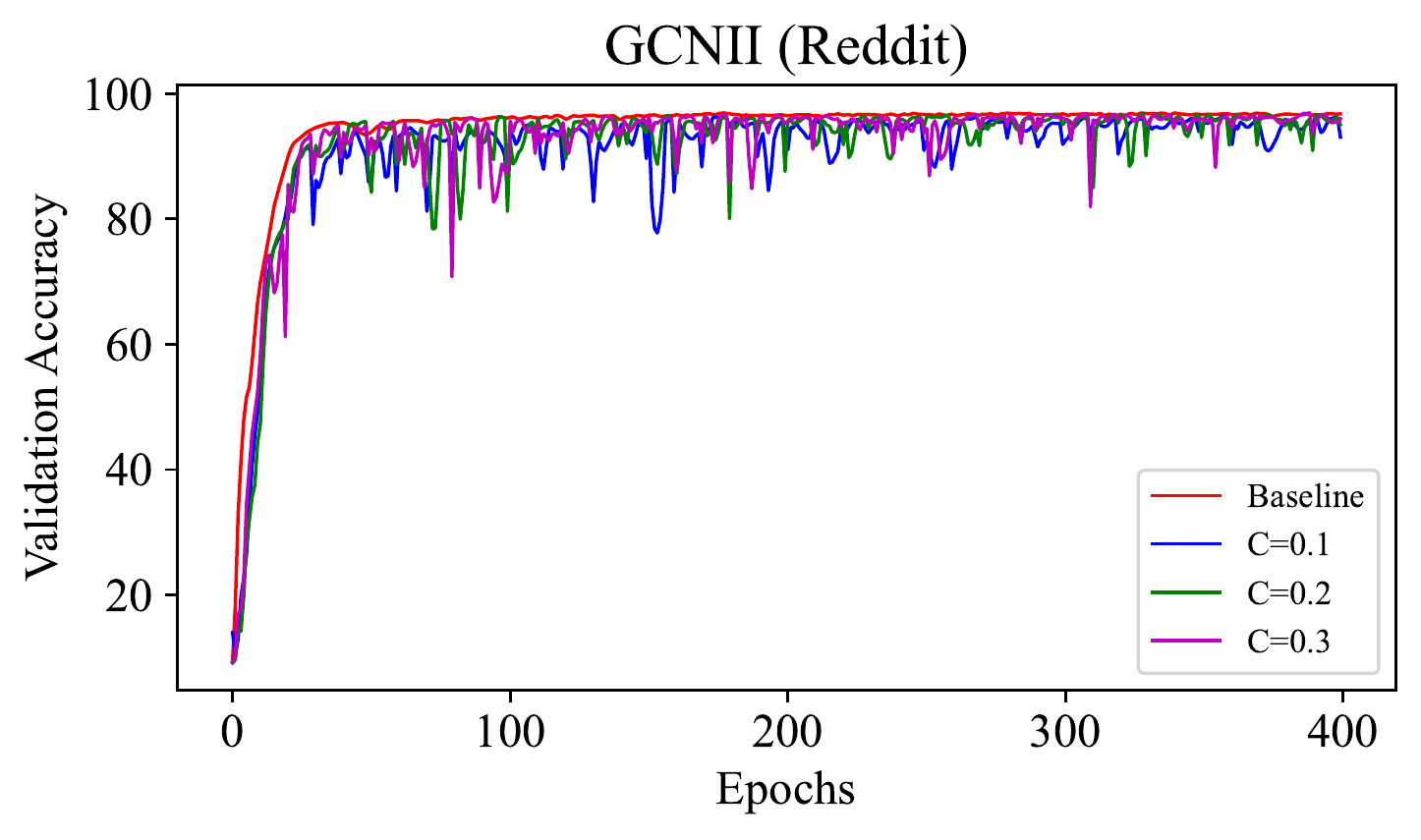}
    \end{subfigure}
    \caption{Learning curves for validation accuracy under different overall budget $C$ on Reddit dataset. Here we disabled the caching and switching mechanism for ablating the effect of $C$.}
    \label{fig: ablation_learning_curve}
\end{figure*}

\end{document}